\documentclass[a4paper]{article}

\usepackage[pages=all, color=black, position={current page.south}, placement=bottom, scale=1, opacity=1, vshift=5mm]{background}
\SetBgContents{
	\tt This work is shared under a \href{https://creativecommons.org/licenses/by-sa/4.0/}{CC BY-SA 4.0 license} unless otherwise noted
}      

\usepackage[margin=1in]{geometry} 

\usepackage{amsmath}
\usepackage{amsthm}
\usepackage{amssymb}
\usepackage{caption}
\usepackage{subcaption}
\usepackage[utf8]{inputenc}
\usepackage{hyperref}
\hypersetup{
	unicode,
	pdfauthor={Susmita Das, Arpita Dutta, Kingshuk Roy, Abir Mondal, Arnab Mukhopadhyay},
	pdftitle={Hate Speech Survey},
	pdfsubject={Hate Speech Survey in Low Resource Languages},
	pdfkeywords={hate, speech, low, resource, language},
	pdfproducer={LaTeX},
	pdfcreator={pdflatex}
}

\usepackage[sort&compress,numbers,square]{natbib}
\usepackage{graphicx, color}
\graphicspath{{fig/}}

\usepackage{algorithm, algpseudocode} 
\usepackage{mathrsfs} 

\title{A Survey on Automatic Online Hate Speech Detection in Low-Resource Languages}
\author{Susmita Das$^1$ \and Arpita Dutta$^2$ \and Kingshuk Roy$^3$ \and Abir Mondal$^4$\and Arnab Mukhopadhyay$^5$}

\date{
	$^1$Bennett University, India \\ \texttt{susmita.das@bennett.edu.in}\\%
	$^2$Techno Main Salt Lake, India \\ \texttt{adutta.7777@gmail.com}\\%
    $^3$Indian Institute of Information Technology \& Management Gwalior, India \\ \texttt{rkingshuk2001@gmail.com}\\%
    $^4$Indian Institute of Engineering Science and Technology, Shibpur, India \\ \texttt{abir22197072@gmail.com}\\%
    $^5$National Institute of Technology, Silchar, India \\ \texttt{arnabmukhopadhyay76@gmail.com}\\%
}

\begin{document}
\maketitle	
\begin{abstract}
The expanding influence of social media platforms over the past decade has impacted the way people communicate. The level of obscurity provided by social media and easy accessibility of the internet has facilitated the spread of hate speech. The terms and expressions related to hate speech gets updated with changing times which poses an obstacle to policy-makers and researchers in case of hate speech identification. With growing number of individuals using their native languages to communicate with each other, hate speech in these low-resource languages are also growing. Although, there is awareness about the English-related approaches, much attention have not been provided to these low-resource languages due to lack of datasets and online available data. This article provides a detailed survey of hate speech detection in low-resource languages around the world with details of available datasets, features utilized and techniques used. This survey further discusses the prevailing surveys, overlapping concepts related to hate speech, research challenges and opportunities.				
\end{abstract}

\section{Introduction}
\label{sec:intro}
In the age of social networking and the exponential growth of social media platforms, people have gained freedom of speech and online anonymity. This has propelled the transmission of hate speech \cite{jahan2023systematic} on multiple social networking platforms and has presently transformed into a global issue. Social media platforms have become a centre for online communications, discussions, opinion exchange and arguments. People's inclination to defend their opinions or ideologies often leads to online debates which occasionally transform into aggressive online behaviour and hate speech. Users often express their animosity in the form of hate speech by berating a certain individual or a group based on their religion, ethnicity, nationality, gender, sex, skin colour or even sexual orientation\cite{shvets2021targets}. Social media companies have implemented their own policies for prevention of hate speech in the cyber-space which often relies on multiple users reporting a certain post or the platform admins regulating the timelines. The evolving social networks depicts increase in prejudiced perceptions, increase speed of hate speech spread\cite{mathew2019spread}, delay in restrictions have posed a great challenge in counteracting against hate speech.\par
Online social networks produce huge amounts of data which makes manual monitoring for hate speech infeasible. This naturally leads to requirement for automatic online hate speech detection techniques. As Natural Language Processing(NLP), Machine Learning(ML) and Deep Learning(DL) approaches make great headway in multiple pragmatic solutions to real-life problems, researchers obtained motivation in analysing different methods for detecting online hate speech. Researchers have assembled multiple large-scale datasets for hate speech, mostly in English language(the most prevalent language worldwide). Although several studies have been organized to deal with hate speech in multiple non-English languages, lack of resources has restricted extensive research. There are various other limitations apart from resource availability which have been acknowledged and require further research. Expanding on the previously available surveys on online hate speech, this paper provides an updated survey on automatic hate speech detection in low-resource languages. In this survey study, we have discussed the following issues associated with online hate speech detection, that collectively constitute our primary contributions:
\begin{enumerate}
    \item Various categories of hate speech and their corresponding background information along with multiple approaches for online hate speech detection are explained elaborately from the existing literature. 
    \item The datasets for hate speech detection are studied well. Moreover, the available datasets for several low resource languages have received special focus.
    \item The multiple cutting-edge natural language processing and deep learning-based techniques used in low-resource languages are analyzed to identify online hate speech.  
    \item The challenges encountered in identifying hate speech along with potential directions for further research have been addressed.
    
\end{enumerate}
\begin{figure}[htbp]
\vspace{-0.3cm}
\centerline{\includegraphics[height=7.82cm,width=0.6\columnwidth]{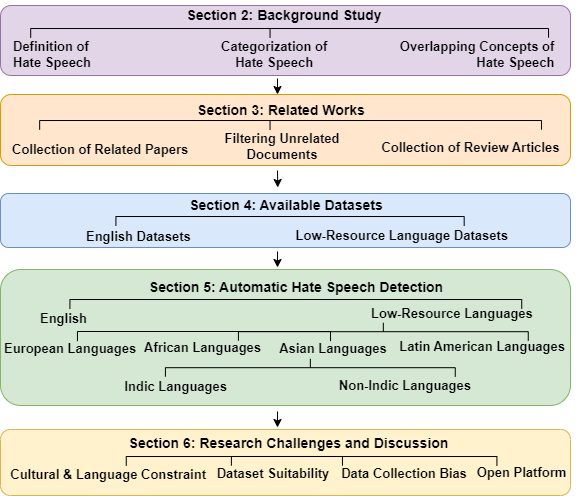}}
\vspace{-0.35cm}
\caption{Survey Overview}
\label{pic:overview}
\end{figure}
The principal objective of this survey is to provide a comprehensive idea about the history of research on hate speech and a detailed account of its application in case of low-resource languages. This survey paper is organized as follows: Section $2$ consists background study detailing the definition of hate speech as explained by different social media giants, categorizing hate speech and the overlapping concepts related to hate speech. Section $3$ explains the search for related works and literature. Section $4$ delves into the datasets available in English and low-resource languages. Section $5$ expounds on automatic hate speech detection, primarily in low-resource languages. These low-resource languages have been presented continent-wise such as European, African, Latin American and Asian to bring clarity. Asian Languages have again been elaborated into Indic Languages and non-Indic languages as India has considerable spoken languages. Section $6$ details the research challenges and other opportunities with Section $7$ consisting of the Conclusion. The overview of this survey article has been depicted in Fig. \ref{pic:overview}.	
\section{Background Study}
\label{sec:pre}
Although hate speech has existed in society for a prolonged time, detection of hate speech from the perspective of computer science is quite a recent development. 
\vspace{-0.3cm}
\subsection{Hate Speech}
\begin{figure}[htbp]
\centerline{\includegraphics[height=6cm,width=0.75\columnwidth]{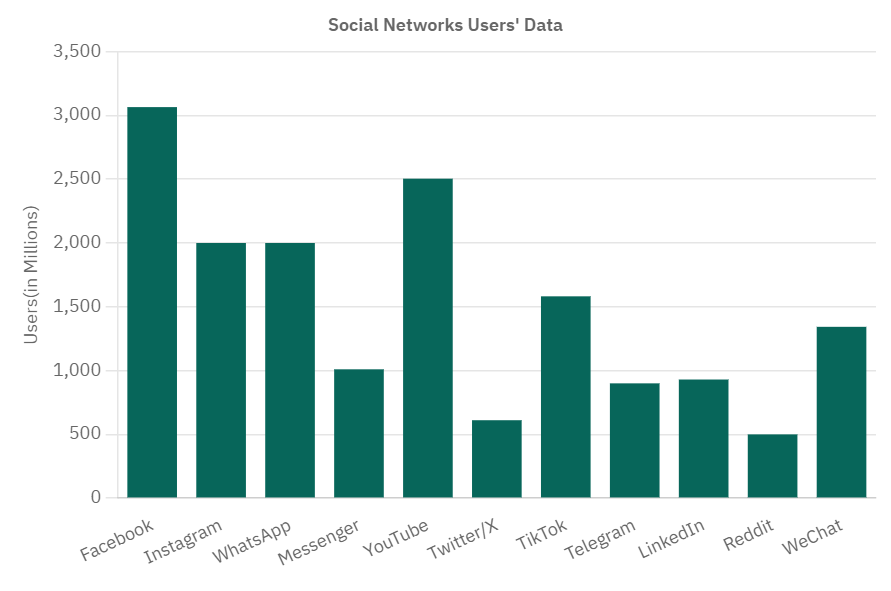}}
\vspace{-0.3cm}
\caption{Monthly Active Users on Different Social Media Platforms\cite{statistasocial}}
\label{pic:users}
\end{figure}

Defining hate speech can be quite difficult as it takes into account complex relationships and viewpoints among different groups of people and the definition is not consistent. 
As of now, the international human rights law does not have a comprehensive definition of hate speech, but UN Strategy and Plan of Action on Hate Speech has defined hate speech as follows:\\
\textit{“any kind of communication in speech, writing or behaviour, that attacks or uses pejorative or discriminatory language with reference to a person or a group on the basis of who they are, in other words, based on their religion, ethnicity, nationality, race, colour, descent, gender or other identity factor.”}\\
Detecting hate speech is challenging as it relies on language variations. In case of online hate speech, multiple social media platforms have given their definition and policies, which are as follows: 
\begin{enumerate}
    \item \textbf{Meta:} Meta, the parent company of multiple social media platforms like Facebook, Instagram etc., has defined hate speech as follows: \textit{"We define hate speech as a direct attack against people – rather than concepts or institutions – on the basis of what we call protected characteristics: race, ethnicity, national origin, disability, religious affiliation, caste, sexual orientation, sex, gender identity and serious disease."}
    \footnote{https://transparency.meta.com/en-gb/policies/community-standards/hate-speech/}
    \item \textbf{YouTube:} YouTube has established its policies to tackle hate speech and harassment. As per official pages of YouTube, hate speech is considered as follows:\textit{"We consider content to be hate speech when it incites hatred or violence against groups based on protected attributes such as age, gender, race, caste, religion, sexual orientation or veteran status."}\footnote{https://www.youtube.com/intl/ALL\_in/howyoutubeworks/our-commitments/standing-up-to-hate/} 
    \item \textbf{Twitter/X:} Twitter, presently known as X, has also provided their hate speech definition: \textit{"You may not attack other people on the basis of race, ethnicity, national origin, caste, sexual orientation, gender, gender identity, religious affiliation, age, disability, or serious disease."}\footnote{https://help.twitter.com/en/rules-and-policies/x-rules}
    \item \textbf{TikTok:} TikTok, a comparatively recent social media platform, has issued their definition: \textit{"Hate speech and hateful behaviour attack, threaten, dehumanise or degrade an individual or group based on their characteristics. These include characteristics like race, ethnicity, national origin, religion, caste, sexual orientation, sex, gender, gender identity, serious disease, disability and immigration status."}\footnote{https://www.tiktok.com/safety/en/countering-hate/}
    \item \textbf{LinkedIn:} LinkedIn is a social media platform focused on employment and professional improvement. As per their policies, they have defined hate speech as: \textit{"Hate speech, symbols, and groups are prohibited on LinkedIn. We remove content that attacks, denigrates, intimidates, dehumanizes, incites or threatens hatred, violence, prejudicial or discriminatory action against individuals or groups because of their actual or perceived race, ethnicity, national origin, caste, gender, gender identity, sexual orientation, religious affiliation, or disability status."}\footnote{https://www.linkedin.com/help/linkedin/answer/a1339812}
\end{enumerate}
As depicted in Fig. \ref{pic:users}, millions of people regularly use multiple social media platforms. It is imperative that these social networking giants implement their policies regarding hate speech to constrain the spread of these discriminatory posts and content.
\vspace{-0.3cm}
\subsection{Categorization of Hate Speech}
The evolving definition of hate speech has included multiple elements under its blanket. Analysis of hate speech contains different categories based on the characteristics of a group that are being targeted. The prevalent categories of hate speech are:
\subsubsection{Racism and Xenophobia:} It is one of the primary categories for hate speech and discrimination. This category includes discrimination of a cluster of people based on their race, country, ethnicity and skin colour. Refugees and immigrants are vilified based on their nationality and the region they belong to. Racial prejudice can be displayed by individuals, communities or institutions. Razavi et al.\cite{razavi2010offensive} show that racism presently also includes cultural aspects.	
\subsubsection{Sexism and Gender Hate Speech:} Sexism refers to stereotyping and animosity against a person based on sex or gender. O'Brien\cite{o2009encyclopedia} has commented that sexism can affect anyone, although the primary targets are females. Gender discrimination is a broader term where bias is displayed against people of different gender identities. Although women are primary targets, this includes non-binary and transgender people. Derogatory remarks against people of other sexual orientations like lesbians, gays and bisexuals are also included in this category.
\subsubsection{Religious Hate Speech:} Another important category of hate speech is discriminating groups of people based on the religions they practice. Religious intolerance leads to stereotyping of certain religious beliefs which often leads to hostility, violence and riots. 
\subsubsection{Ableism:} Negative attitudes are often displayed against people with disabilities. Friedman et al.\cite{friedman2017defining} have proposed these disabilities may include physical, sensory as well as cognitive disabilities. Sometimes this may include discrimination against people with certain diseases. This category of hate speech manifests from the belief that people with disabilities are somehow inferior.\par
\begin{table}[!ht]
\caption{Analysis of Hate Speech Definitions in Different Platforms}
\label{table:def}
\centering
\resizebox{\textwidth}{!}{%
\begin{tabular}{cccccc}
\hline
          & Race/Ethnicity & \begin{tabular}[c]{@{}c@{}}Sexism/Gender \end{tabular} & Religion & Ableism & Others                                     \\
\hline\\
Meta      & Yes            & Yes                                                                    & Yes      & Yes     & Caste, Serious Disease                     \\
YouTube   & Yes            & Yes                                                                    & Yes      & No      & Age, Veteran Status                        \\ 
Twitter/X & Yes            & Yes                                                                    & Yes      & Yes     & Caste, Serious Disease                     \\
Tiktok    & Yes            & Yes                                                                    & Yes      & Yes     & Caste, Serious Disease, Immigration Status\\
LinkedIn  & Yes      &   Yes     & Yes    & Yes   & Caste\\
\hline
\end{tabular}%
}
\end{table}
As observed in Table \ref{table:def}, social media platforms have defined hate speech as discrimination against specific groups. Most of these definitions have considered religion, race, gender discrimination, colour, ethnicity, disability. Some other criteria like serious disease, age, immigration status etc, have also been included in hate speech definition. 
\subsection{Correlation and Overlapping Concepts of Hate Speech}
The categories of hate speech are correlated and have interrelationships among them. The extensive usage of social media have allowed the spread of abusive language online. People are often attacked on individually and harassed on the online social networks. Some concepts of these persecutions are:
\subsubsection{Hate:} While hate speech is stereotyping and focusing on a particular group, hate is generally displaying aggressive behaviour for no particular reason.
\subsubsection{Discrimination:} Discrimination is identifying a certain group or community and making unfair distinctions, prejudiced treatment. Hate speech is basically discrimination, but online or verbal in nature.
\subsubsection{Cyberbullying:} Cyberbullying is targeted harassment of a certain individual on social media platforms. Dinakar et al.\cite{dinakar2012common}, Smith et al.\cite{smith2008cyberbullying} have explained that much like traditional bullying, cyberbullying involves repeated aggression towards the same individual or group of individuals.
\begin{table}[!ht]
\caption{Hate Speech Categorization}
\label{table:category}
\centering
\resizebox{\textwidth}{!}{%
\begin{tabular}{lcccccc}
\hline
   & \begin{tabular}[c]{@{}c@{}}Race/\\ Ethnicity\end{tabular} & \begin{tabular}[c]{@{}c@{}}Sexism/Gender\end{tabular} & Religion & Ableism & Cyberbullying & Abusive \\
   \hline
\begin{tabular}[l]{@{}l@{}}1. "He is a good  person."\end{tabular}                                                                & No                                                        & No                                                                     & No       & No      & No            & No      \\
\hline
\begin{tabular}[l]{@{}l@{}}2. "May be he is a  cheapskate."\end{tabular}                                                                 & No                                                        & No                                                                     & No       & No      & No            & Yes     \\
\hline
\begin{tabular}[l]{@{}l@{}}3. "Why doesn't he realize \\ he is a chunker. These clothes \\ are not for him"\end{tabular}         & No                                                        & No                                                                     & No       & No      & Yes           & Yes     \\
\hline
\begin{tabular}[l]{@{}l@{}}4. "This person with \\ \textless{}insert skin colour\textgreater  is \\ so dirty"\end{tabular}         & Yes                                                       & No                                                                     & No       & No      & Yes           & Yes     \\
\hline
\begin{tabular}[l]{@{}l@{}}5. "These two men are obnoxious. \\ I don't want to see these homos"\end{tabular}                     & No                                                        & Yes                                                                    & No       & No      & Yes           & Yes     \\
\hline
\begin{tabular}[l]{@{}l@{}}6. "Never seen a   \textless{}insert religion\textgreater \\who is a pansy ;) ."\end{tabular} & No                                                        & Yes                                                                    & Yes      & No      & Yes           & Yes     \\
\hline
\begin{tabular}[l]{@{}l@{}}7. "He is a moron. People call  \\him THE DUMB."\end{tabular}                                            & No                                                        & No                                                                     & No       & Yes     & Yes           & Yes     \\
\hline
\begin{tabular}[l]{@{}l@{}}8. "People from \textless{}insert country\textgreater\\ are not good."\end{tabular}                    & Yes                                                       & No                                                                     & No       & No      & No            & Yes    \\
\hline
\begin{tabular}[l]{@{}l@{}}9. "People from \textless{}insert religion\textgreater\\ should not be allowed inside."\end{tabular}                    & No                                                       & No                                                                     & No       & Yes      & No            & No    \\
\hline
\end{tabular}%
}
\end{table}
\subsubsection{Abusive Language:} Abusive language refers to offensive, rude and derogatory language. Abusive language includes profanities, demeaning, slanderous and swear words which evokes inferior sentiment in other individuals. Hate speech can be considered as a variety of abusive language.\par
\begin{figure}[htbp]
\centerline{\includegraphics[height=5cm,width=0.75\columnwidth]{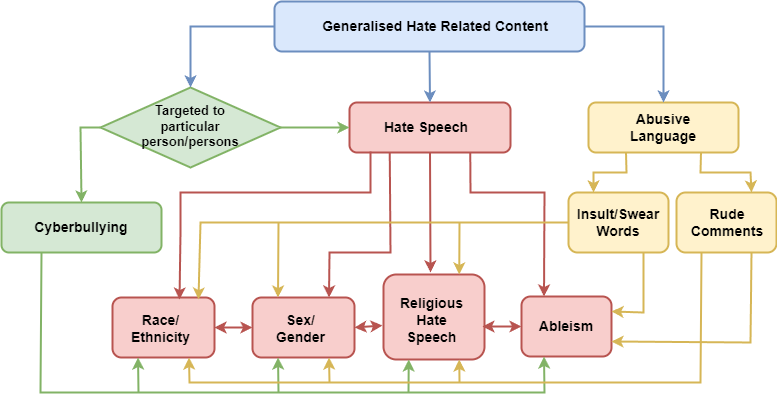}}
\vspace{-0.3cm}
\caption{Relation between Hate Speech and Extended Concepts}
\label{pic:category}
\end{figure}
"Cyberbullying", "Abusive Language", "Hate Speech" have a lot of similarities among themselves and their concepts often overlap with each other. Recognizing and classifying online contents accordingly can become challenging. Racial slurs, sexist terms etc. can be used to bully a person online. In this case, a particular post can be classified as both hate speech and cyberbullying or abusive. There are instances where, the message is a case of cyberbullying but cannot be regarded as hate speech. As shown in Table \ref{table:category}, a certain post or message can be designated under multiple categories of hate speech. As for example, in sentence-4 of Table \ref{table:category}, (\textit{"This person with
<insert skin colour> is so dirty"}) is categorised as abusive, cyberbullying and at the same time racist hate speech. Similarly, in sentence-6, (\textit{"Never seen a \textless{}insert religion\textgreater  who is a pansy ;) ."}) contains abusive term (\textit{"pansy"}) and is categorised as both religious hate speech and cyberbullying. Sentence-2 contains abusive word \textit{"cheapskate"}, but it may not be categorised as cyberbullying, due to the fact that the writer of the post is just wondering not targeting the person. Again, sentence-9, (\textit{"People from <insert religion>
should not be allowed inside."}), does not contain any abusive language or cannot be categorised as cyberbullying. This sentence is definitely discriminating a group of people based on religion which is hate speech. These examples demonstrates that cyberbullying and abusive language can be potential determinants of hate speech, but presence or non-presence of these components does not determine whether a certain message is hate speech. As in Fig. \ref{pic:category}, hate speech and the other related or adjacent concepts have intersections in their relationships with each other.\par 
\begin{figure}
     \centering
     \begin{subfigure}[htbp]{0.47\columnwidth}
         \centering
         \includegraphics[width=7cm,height=5cm]{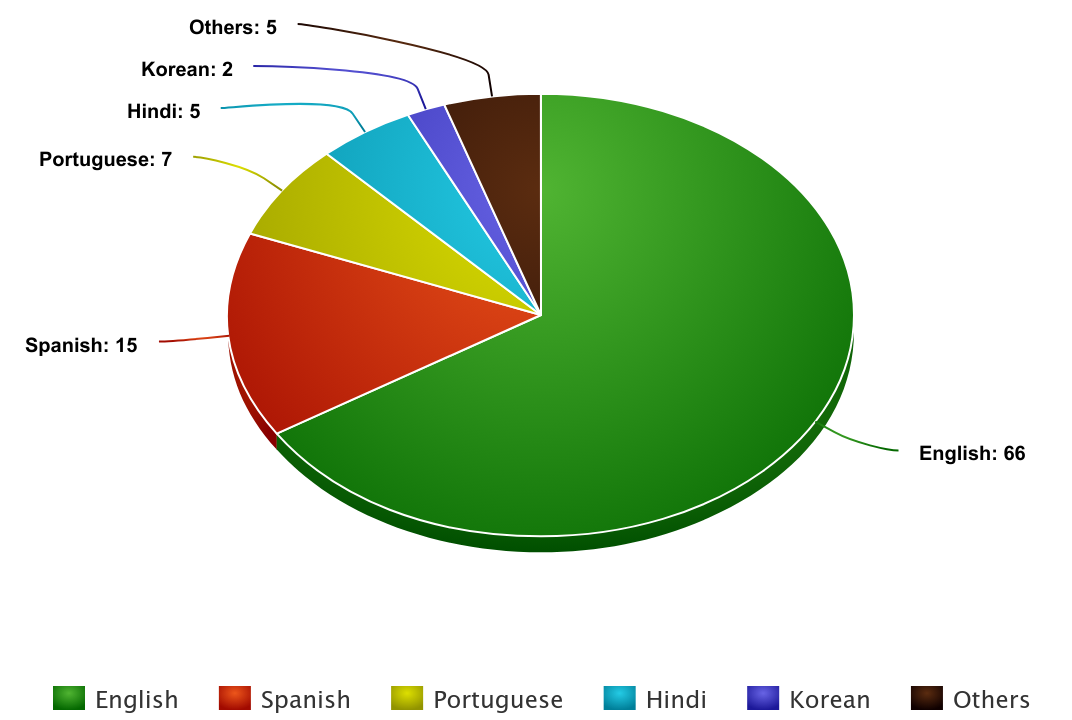}
         \caption{Content Percentage of Frequently Used Languages on YouTube\cite{wikipediayou}}
         \label{fig:YouTube_3D}
     \end{subfigure}
     \hfill
     \begin{subfigure}[htbp]{0.47\columnwidth}
         \centering
         \includegraphics[width=7cm, height=5cm]{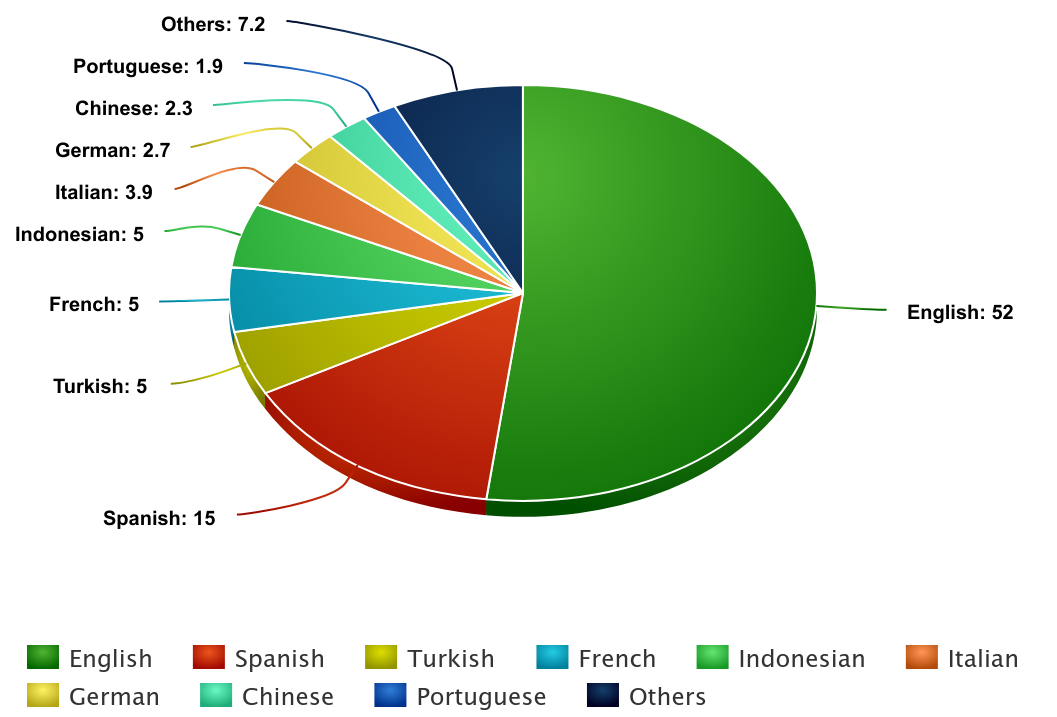}
         \caption{Content Percentage of Frequently Used Languages on Facebook\cite{linkedfb}}
         \label{fig:FB_3D}
     \end{subfigure}
     \hfill
     \begin{subfigure}[htbp]{0.47\columnwidth}
         \centering
         \includegraphics[width=7cm, height=5cm]{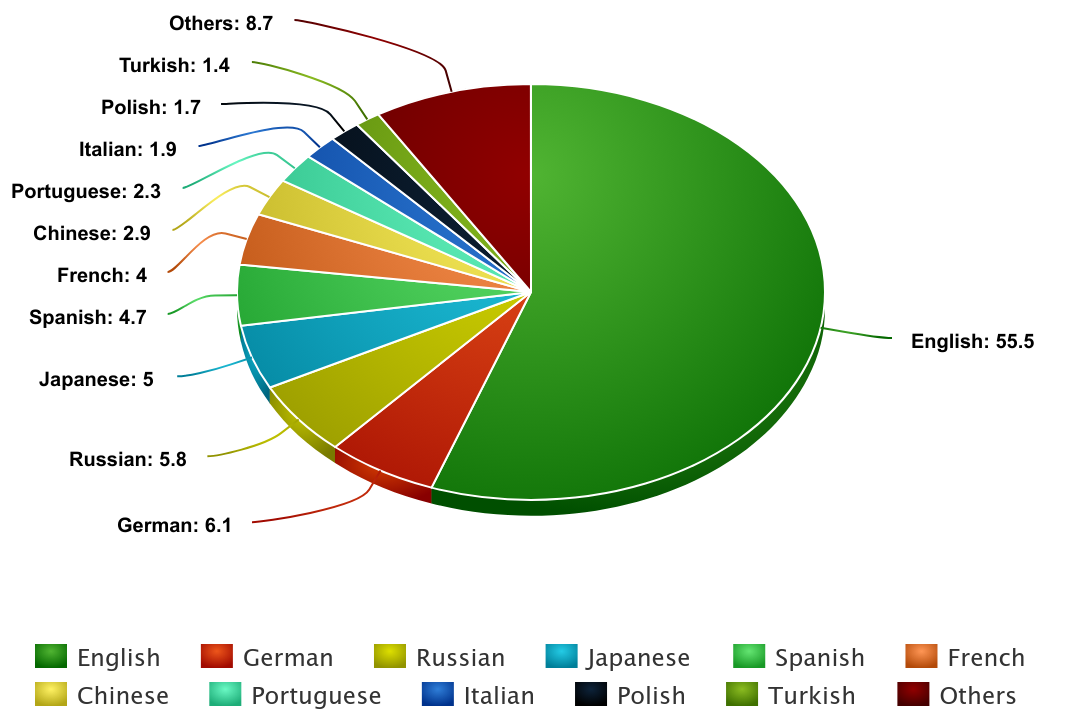}
         \caption{Content Percentage of Frequently Used Languages on Twitter/X\cite{semioX}}
         \label{fig:X_3D}
     \end{subfigure}
     \hfill
     \begin{subfigure}[htbp]{0.47\columnwidth}
         \centering
         \includegraphics[width=7cm, height=5cm]{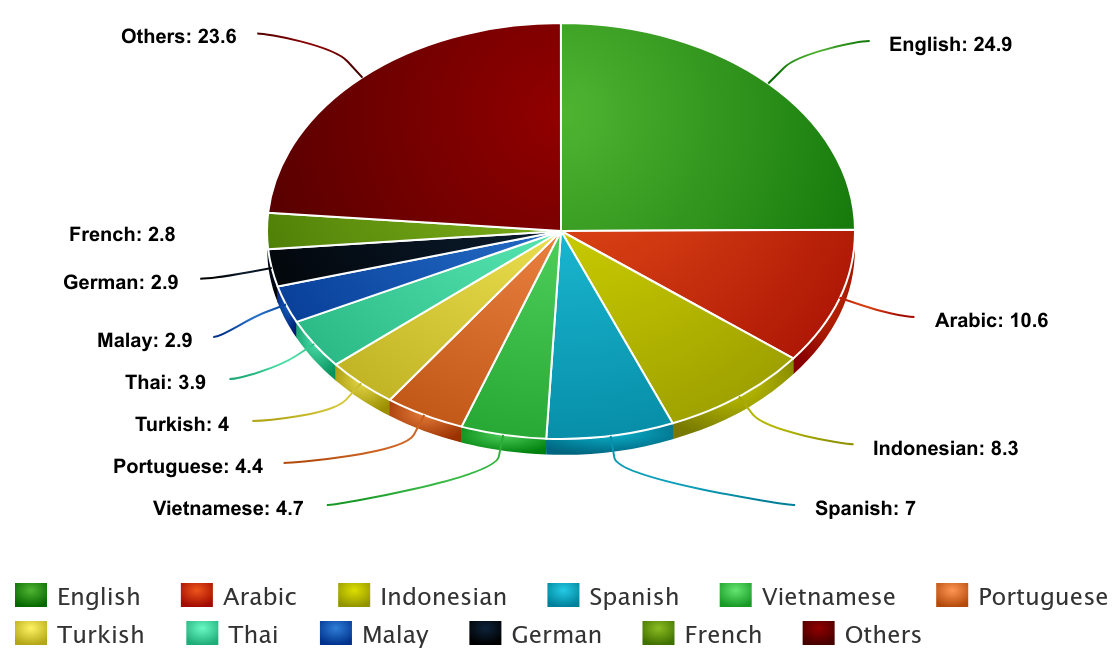}
         \caption{Content Moderator Languages on TikTok\cite{Tiklang}}
         \label{fig:Tik_3D}
     \end{subfigure}
     \hfill
        \caption{Content Percentage of Languages on Various Social Media Platforms}
        \label{fig:Social}
\end{figure}
Due to complexities of language formation and semantics, annotation of dataset for hate speech can be ambiguous. Although abusive language has been used as a blanket term for many categories of hate speech. Van Hee et al.\cite{van2015guidelines}  has considered racist and sexist remarks as part of offensive language, while Davidson et al.\cite{davidson2017automated} clearly distinguishes both. In case of annotations of online posts, while some posts are considered as hate speech by one research\cite{waseem2016hateful}, those same posts are considered as abusive language in another study\cite{nobata2016abusive}. Thus, complex language norms as well as ambiguous dataset annotation, both contribute to complications in hate speech detection.
\vspace{-0.2cm}
\section{Related Works}
Hate speech has been explored in multiple research topics. There are multiple studies regarding hate speech in politics, law, journalism, human behaviour etc. When it comes to computer science, studies on hate speech have been a comparatively recent trend. A substantial amount of resources are available in public domain encouraging researchers to study online hate speech. As English is the prevailing language for information exchange on the internet, most resources available are predominantly in English.
Apart from English, account holders on various social media platforms have started to use their native languages for communication purposes. These native languages can be considered online low-resource languages, as considerably fewer resources can be obtained for these languages. As depicted in Fig. \ref{fig:Social}, it has been observed that English is the dominant language used in all social media platforms. After English, it is discerned that there is an abrupt decrease in the content percentage of other languages across all platforms. While Spanish is the second-most used language on both YouTube and Facebook as shown in Fig. \ref{fig:YouTube_3D} and Fig. \ref{fig:FB_3D} respectively, German is the second-most used language on Twitter/X as portrayed in Fig. \ref{fig:X_3D}, while Arabic is the second-most used language by moderators on TikTok as shown in Fig. \ref{fig:Tik_3D}.\par
The number of research articles and available datasets is not vast. In the case of low-resource languages, the frequency of articles is quite sparse. However, we accumulated a considerable number of scholarly articles from multiple digital libraries and websites.
\vspace{-0.3cm}
\begin{table}[htbp]
\caption{Keywords used for Relevant Document Search}
\label{table:keyword}
\centering
\resizebox{\textwidth}{!}{%
\begin{tabular}{ccc}
\hline
\textbf{Hate Speech related Keywords}   & \textbf{Dataset Related Keywords}     & \textbf{Language Related Keywords}     \\
\hline
\begin{tabular}[c]{@{}c@{}}Hate Speech Detection, Hate Speech Survey,\\ Hate Speech Detection Systematic Review,\end{tabular}             & Hate Speech Dataset                                                                              & \begin{tabular}[c]{@{}c@{}}Spanish, Dutch, Danish, Italian,\\ French, German, Portuguese, Greek\end{tabular} \\
\begin{tabular}[c]{@{}c@{}}Cyberbullying, Cyberbullying Detection,\\ Cyberbullying Survey, Trolling\end{tabular}                                    & \begin{tabular}[c]{@{}c@{}}Offensive Language Dataset,\\ Abusive Language Dataset\end{tabular}   & \begin{tabular}[c]{@{}c@{}}Turkish, Indonesian, Korean, \\ Mandarin, Chinese, Japanese\end{tabular}  \\
\begin{tabular}[c]{@{}c@{}}Racism, Racism on social media, Racist speech,\\ Ableism, Disability\end{tabular}                                        & Multilingual Dataset                                                                             & \begin{tabular}[c]{@{}c@{}}Arabic, Swahili, Nepali,\\ Vietnamese, Bengali, Urdu\end{tabular}                 \\
\begin{tabular}[c]{@{}c@{}}Sexism, Sexism on Social Media, Sexist Speech,\\ Misogyny, Gender Discrimination, Homophobia,\\ Transgender\end{tabular} & \begin{tabular}[c]{@{}c@{}}Low-Resource Dataset,\\ Multimodal Hate Speech\\ Dataset\end{tabular} & \begin{tabular}[c]{@{}c@{}}Hindi, Marathi, Assamese\\ Telugu, Kannada, Telugu\end{tabular}                   \\
Religious Hate Speech, Religious Discrimination                                                                                                     & Benchmark Dataset                                                                                & \begin{tabular}[c]{@{}c@{}}Hinglish, Bengali code-mixed,\\ Hindi code-mixed\end{tabular}                     \\
\begin{tabular}[c]{@{}c@{}}Abusive Language, Offensive Language,\\ Swear Words, Toxic Comments, Insults\end{tabular}                                & Code-Mixed Dataset                                                                               & \begin{tabular}[c]{@{}c@{}}Indic Language, Dravidian\\ Language, Asian Language\end{tabular}                
\end{tabular}%
}
\end{table}
\vspace{-0.3cm}
\subsection{Method of collecting related papers}
\begin{figure}[htbp]
\centerline{\includegraphics[height=7cm,width=0.75\columnwidth]{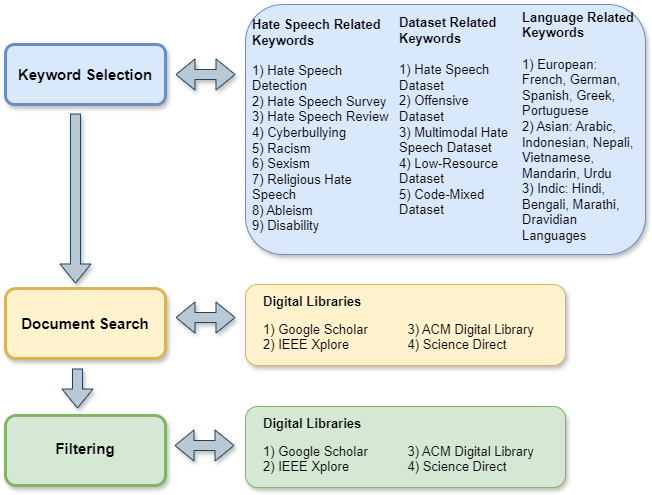}}
\vspace{-0.3cm}
\caption{Collection of Relevant Documents}
\label{pic:collection}
\end{figure}

While surveying hate speech detection in online platforms, we studied the previous reviews and implementation papers. The process for collection of relevant documents have been depicted in Fig. \ref{pic:collection}. 
\subsubsection{Keywords Determination:} For retrieval of relevant research articles, we accumulated information from online literature databases and digital libraries like Google Scholar, IEEE Xplore, ACM Digital Library, Science Direct etc. The concept of hate speech also encompasses cyberbullying, abusive and offensive language. During keyword selection along with 'Hate Speech', we included 'Racism', 'Sexism', 'Religious Discrimination', 'Homophobia', 'Cyberbullying', 'Abusive', 'Offensive', 'Toxic' etc. We paid particular attention to hate speech in low-resource languages. We included multiple search terms regarding the low-resource languages to retrieve a significant amount of documents to explore. An extensive list of keywords used for fetching documents have been depicted in Table \ref{table:keyword}.
\subsubsection{Related Document Search:} During collection of literary documents, the research based on computer science has been kept in focus. The title of the paper, author details, research details from the abstract, datasets used, methodology implemented, the results obtained and full link of the research article are extracted and summarised. Initially, we were attentive towards accumulating papers on hate speech detection and no language restrictions were imposed. In the primary stage, the papers gathered were mostly research on the English language. We assembled the surveys and review papers on hate speech in computer science domain. These reviews and survey papers summarise the datasets, corpora and methodologies used for hate speech detection. In later stages, our focus shifted to assembling information on low-resource languages. Our search became more specific on gathering low-resource datasets, and automatic hate speech detection articles on multiple low-resource languages around the world as well as in India. We have gathered documents from 2005 but our concentration has been primarily on research articles between 2017 to current time.
\subsection{Filtering of Unrelated Documents}
As hate speech has been an interdisciplinary research area, we had to be particular about the scholarly articles that were considered relevant to our survey. A lot of information obtained were either unrelated to hate speech, different hate speech categories or immaterial to computer science. Duplication of documents during the collection was managed. In case of review or survey papers, articles were carefully scrutinised. While some survey papers are based on the methods used for automatic hate speech detection, some are based on hate speech datasets and corpus. Research papers with indeterminate information were excluded.  
\subsection{Literature Review of Survey/Review Articles}
Surveys and review articles are eminent means of gaining knowledge and resources regarding a particular research field. We collected review articles regarding hate speech and abusive language datasets and hate speech detection techniques.
Fortuna and Nunes\cite{fortuna2018survey} presented a systematic review on the evolution of automatic hate speech detection in the textual field. This review has provided a critical overview of the theoretical concepts and the available practical datasets. Alkomah and Ma\cite{alkomah2022literature} have introduced a systematic approach of providing different available hate speech dataset details and summarisation of multiple textual hate speech detection techniques. Mansur et al.\cite{mansur2023twitter} proposed a systematic review of hate speech detection techniques specifically based on Twitter datasets and resources.\par
Review articles on low-resource languages are not easy to come across, as the amount of research regarding these languages is not abundant. 
Al-Hassan and Al-Dossari\cite{al2019detection} have provided a summary on the multilingual facet of hate speech identification. Their survey has included English hate speech as well as Arabic anti-social behaviours. Jahan and Oussalah\cite{jahan2023systematic} have presented a systematic review of the multilingual datasets available in hate speech and an overview of the deep learning models being used. Del et al.\cite{del2023sentiment} used datasets HatEval2, PAN (CLEF2021), ParlSpeech dataset, media news, and college media. The systematic review discusses lexicon-based and machine-learning techniques, including BERT (Bidirectional Encoder Representations from Transformers) and its Spanish-trained model BETO, NLTK, Stanford NLP, Word2Vec, and Freeling for various NLP tasks. It discusses the state of the art, identifies gaps, and highlights challenges in sentiment analysis for Spanish language content related to politics and hate speech. Dhanya and Balakrishnan\cite{dhanya2021hate} have organised a hate speech detection survey in different Asian languages. This survey includes research on low-resource Asian languages such as Indonesian, Arabic, Roman-Urdu, Sinhala, Nepali, Hindi, Bengali, Dravidian languages etc. Mahmud et al.\cite{mahmud2023cyberbullying} provides an extensive review of cyberbullying detection methods being used in the case of low-resource languages. Chhabra et al.\cite{chhabra2023literature} presented a survey of hate speech detection techniques in multiple languages as well as multiple modalities. Pamungkas et al.\cite{pamungkas2023towards} introduces a survey of abusive language detection methods across multiple domains in multiple languages.\par
Recently, an increasing number of people are using their native languages to communicate on social media platforms. Accordingly, the number of surveys on hate speech, cyberbullying and abusive words in low-resource languages has increased in the past few years. Our motivation has been to provide an up-to-date literature review highlighting the methods used for hate speech detection in low-resource languages along with the resources and datasets available in these languages.  
\section{Available Datasets}
Several valuable resources have been identified during the course of this  review. While most of the resources and datasets are open source projects, there are few resources for which permission is needed from the authors and creators of the datasets. Most of the publicly accessible datasets are present in GitHub, while a few of them are obtainable from different other repositories. Majority of the datasets have been compiled from public resources available on multiple social media platforms. The prevalent social media platform for data collection is observed to be Twitter/X, while Facebook and Reddit are dataset sources. A substantial number of studies have also considered YouTube, Gab etc. English is dominant language in which a major part of the studies have been conducted. Research have been carried out on other European languages like Spanish, French, German etc. Recently, studies on Asian languages like Arabic, Korean and multiple Indic languages are gaining attention.
\subsection{English Language Datasets:}
Regarding datasets, there are abundant public resources available in English. Depending on the requirements of the research, authors have collected posts from social media platforms, annotated them for binary classification or multi-label classification. Details of a few widely used English datasets have been provided below.
\subsubsection{Waseem Dataset:}\footnote{https://github.com/ZeerakW/hatespeech} Waseem et al.\cite{waseem2016hateful} have gathered this dataset from Twitter consisting of 16k tweets. In this 3383 tweets were sexist in nature and posted by 613 users, 1972 had racist content conveyed by 9 users, and 11,559 had neither sent by 614 users. After annotating the data themselves, it was verified by third-party annotators.
\subsubsection{Davidson Dataset:}\footnote{https://github.com/t-davidson/hate-s
peech-and-offensive-language} Davidson et al.\cite{davidson2017automated} constructed this Twitter dataset using lexicon from Hatebase.org, consisting of words and phrases with hate speech. It consists of approximately 25k English tweets which were annotated using crowd-sourcing. Each tweet was annotated into three categories: hate speech, only offensive but not hate speech and neither. In this dataset 5.77\% tweets were labelled as Hate speech, 77.43\% labelled as Offensive and 16.80\% as neither. 
\subsubsection{Qian Dataset:}\footnote{https://github.com/jing-qian/A-Bench
mark-Dataset-for-Learning-to-Intervene-i
n-Online-Hate-Speech} Qian et al.\cite{qian2019benchmark} accumulated their English language dataset from both Reddit and Gab which were conversational threads to preserve the context of hate speech. They crowd-sourced the intervention responses from workers at Mechanical Turk. The Reddit dataset consists of 5020 conversations incorporating 22324 comments annotated as hate speech or
non-hate. Major part of the dataset with 76.6\% conversations consisted of hate speech, while the others were non-hate. 7641 unique intervention responses were collected. The Gab dataset consists of 11,825 conversations and total of 33776 posts. Almost the whole dataset with 94.5\% conversations consists of hate speech. 21747 distinctive intervention responses were accumulated in Gab dataset.
\subsubsection{Zhang Dataset:}\footnote{https://github.com/ziqizhang/data\#hate} Zhang et al.\cite{zhang2018detecting} collected their dataset from Twitter consisting of 2435 tweets. These accumulated tweets are specific to refugees, immigrants and muslims. In this dataset, 414 has been annotated as hate while 2021 has been labelled as non-hate.
\subsubsection{ETHOS Dataset:}\footnote{https://github.com/intelligence-csd-auth-gr/Ethos-Hate-Speech-Dataset} Mollas et al.\cite{mollas2022ethos} assembled this dataset based on Reddit and YouTube comments and validated the annotations from Figure-Eight crowd-sourcing platform. There are two-variants of this dataset: binary variation with 998 samples and multi-label variation with 433 samples. The binary variation consists of label 'isHate' and not, while the multi-label variation consists of 8 labels: ‘violence’, ‘directed\_vs \_generalised’, ‘gender’, ‘sexual\_orientation’,  ‘disability’, ‘race’, ‘national\_origin’, ‘religion’. For binary, 44.39\% comments consists hate speech and in 55.61\% comments there is absence of hate speech. For multi-label,19.41\% comments are for gender, 16.48\% are related to sexual orientation, 11.96\% for disability, 17.16\% for race, 16.70\% connected to national origin and lastly 18.28\% for religion. 
\subsubsection{HateXplain Dataset:}\footnote{https://github.com/punyajoy/HateXplain} Mathhew et al.\cite{mathew2021hatexplain} have compiled this dataset with posts in English from Twitter and Gab which annotates hate speech based on both word and phrase level. This dataset has been annotated by Amazon Mechanical Turk workers in three labels: 'hate', 'offensive' and 'normal'. The annotators determined the targeted communities and the parts of text in these comments which explain their classification. There are a total of 20148 samples in this dataset with 9055 from Twitter and 11093 from Gab. In the dataset, 29.457\% of total samples are labelled as hate (7.81\% of total Twitter samples and 57.72\% of total Gab samples), 27.19\% as offensive (25.71\% of total Twitter samples and 34.81\% of total Gab samples), 38.78\% as normal (63.72\% of total Twitter samples and 22.57\% of total Gab samples). 4.56\% are labelled as undecided where each annotator chose different labels. 
\subsection{Low-Resource Languages Datasets:} Compilation of datasets in low-resource languages is quite challenging. The publicly available data on different social media platforms is sparse. Regarding openly accessible resources for data selection, researchers also have to consider transliteration and code-mixed languages. Several low-resource datasets are purely in a single language, other low-resource datasets are multilingual. Details of both monolingual and multilingual datasets containing dataset size, dataset link, annotation classes have been provided below.
\subsubsection{Monolingual Datasets:} Monolingual datasets have been collected in multiple low-resource languages across the world such as Greek, Italian Twitter Corpus, Portuguese dataset, Indic languages. Details of these datasets have been provided below, while summary of multiple other datasets have been depicted in Table \ref{table:non-E_data}.\\ 
\begin{table}[!ht]
\caption{Monolingual Datasets in Low-resource Languages}
\label{table:non-E_data}
\centering
\resizebox{\textwidth}{!}{%
\begin{tabular}{|c|c|c|c|c|c|}
\hline
\textbf{Name}  & \textbf{Platform} & \textbf{Language} & \textbf{Size}    & \textbf{Labels}     & \textbf{Reference}                                         \\ \hline 

\begin{tabular}[c]{@{}c@{}}Offensive Greek\\ Tweet Dataset\\ (OGTD) - 2020\footnote{https://zpitenis.com/ogtd}\end{tabular} & Twitter           & Greek             &  5000  & \begin{tabular}[c]{@{}c@{}}Offensive,\\ Not Offensive, Spam\end{tabular}   & Pitenis et al.\cite{pitenis2020offensive} \\ \hline

\begin{tabular}[c]{@{}c@{}}Portuguese Hate Speech\\  Dataset - 2019\footnote{https://github.com/paulafortuna/Port uguese-Hate-Speech-Dataset}\end{tabular}     & Twitter           & Portuguese           & 5668 & \begin{tabular}[c]{@{}c@{}}Binary(HS, Not-HS),\\ Hierarchical(Sexism,\\ Racism etc.)\end{tabular}  & Fortuna et al.\cite{fortuna2019hierarchically}      \\ \hline

\begin{tabular}[c]{@{}c@{}}Italian Twitter\\ Corpus - 2017\end{tabular}       & Twitter           & Italian           &  1828 & Hate, Non-Hate & Poletto et al.\cite{poletto2017hate}      \\ \hline

\begin{tabular}[c]{@{}c@{}}Italian Twitter\\ Corpus -2018\footnote{https://github.com/msang/hate-speech-corpus}\end{tabular}       & Twitter           & Italian           &  1827 & Immigrants, Not &\begin{tabular}[c]{@{}c@{}} Sanguinetti et al.\\\cite{sanguinetti2018italian} \end{tabular}     \\ \hline

\begin{tabular}[c]{@{}c@{}}PolicycorpusXL\\ 2021\end{tabular}       & Twitter           & Italian           &  7000 & Hate, Normal & Celli et al.\cite{celli2021policycorpus}      \\ \hline

\begin{tabular}[c]{@{}c@{}}Anti-Social\\ Behaviour in Online\\ Communication\\-2018\footnote{https://goo.gl/27EVbU} \end{tabular}   & YouTube   & Arabic   &   15050 &  \begin{tabular}[c]{@{}c@{}}Offensive, Not Offensive \end{tabular}  & Alakrot et al.\cite{alakrot2018dataset}   \\ \hline

\begin{tabular}[c]{@{}c@{}}Arabic Religious\\ Hate Speech\\-2018\footnote{https://github.com/nuhaalbadi/Arabic hatespeech} \end{tabular}   & Twitter   & Arabic   &   16914 &  \begin{tabular}[c]{@{}c@{}}Hate, Not Hate \end{tabular}  & Albadi et al.\cite{albadi2018they}   \\ \hline

\begin{tabular}[c]{@{}c@{}}European Refugee\\ Crisis-2017\footnote{https://github.com/UCSM-DUE/
IWG\_hatespeech\_public
}\end{tabular} & Twitter & German & 469 & \begin{tabular}[c]{@{}c@{}}Hate Speech,\\ Toxicity, Threat\end{tabular}  & Ross et al.\cite{ross2017measuring}   \\ \hline

\begin{tabular}[c]{@{}c@{}}DeTox\footnote{https://github.com/ hdaSprachtechnologie/detox}\end{tabular} & Twitter & German & 10278 & \begin{tabular}[c]{@{}c@{}}Anti-Refugee Hate,\\ Not Hate \end{tabular}  & Demus et al.\cite{demus2022detox}   \\ \hline

\begin{tabular}[c]{@{}c@{}}Indonesian Hate\\ Speech Dataset-2017\footnote{https://github.com/ialfina/id-hatespeech-detection}\end{tabular} & Twitter & Indonesian & 520 & Hate, Not Hate  & Alfina et al.\cite{alfina2017hate}   \\ \hline

\begin{tabular}[c]{@{}c@{}} SOLD\\
2024\footnote{https://huggingface.co/datasets/sinhala-nlp/SOLD} \end{tabular}   & Twitter  & Sinhala   &   10000 &  Hate, Abusive  & Ranasinghe et al.\cite{ranasinghe2024sold}   \\ \hline

\begin{tabular}[c]{@{}c@{}} Hinglish Offensive\\
Tweet(HOT)\\Dataset-2018\footnote{https://github.com/pmathur5k10/Hinglish-Offensive-Text-Classification} \end{tabular}   & Twitter  & Hindi   &   3189 &  Hate, Abusive  & Mathur et al.\cite{mathur2018did}   \\ \hline

\begin{tabular}[c]{@{}c@{}} Bengali Hate\\
Speech Dataset-2020\footnote{https://github.com/rezacsedu/Bengali-Hate-Speech-Dataset} \end{tabular}   & \begin{tabular}[c]{@{}c@{}}YouTube,\\Facebook,\\News articles\end{tabular}  & Bengali   &   10000 &  \begin{tabular}[c]{@{}c@{}}Hate(different\\categories), Abusive \end{tabular} & Karim et al.\cite{karim2020classification}   \\ \hline

\begin{tabular}[c]{@{}c@{}} L3Cube-MahaHate\\
Dataset-2022\footnote{https://github.com/l3cube-pune/MarathiNLP} \end{tabular}   & Twitter  & Marathi   &    25000 &  \begin{tabular}[c]{@{}c@{}}Hate, Offensive,\\Profane, Not \end{tabular} & Velankar et al.\cite{velankar2022l3cube}   \\ \hline
\end{tabular}%
}
\end{table}
\\
\textbf{1. Coltekin Dataset:}\footnote{https://coltekin.github.io/offensive-turkish/} Coltekin\cite{ccoltekin2020corpus} have compiled this monolingual dataset consisting of 36232 tweets in Turkish language from the Twitter which have been sampled randomly. A major portion, 92.1\% of the tweets are from unique users. A hierarchical annotation process has been followed during labelling of tweets. In the top level, the tweets are annotated as offensive or non-offensive. The offensive tweets are further identified with targeted communities and whether these are targeting individuals, groups or others. In this dataset, 80.6\% are non-offensive tweets, while 19.4\% are offensive tweets. From these offensive tweets, 21.18\% are not targeted to any community, 25.45\% are targeted to groups, 48.02\% are targeted towards individuals and 5.33\% towards others.\\ 
\\
\textbf{2. Beyhan Dataset:}\footnote{https://github.com/verimsu/
Turkish-HS-Dataset} Beyhan et al.\cite{beyhan2022turkish} accumulated the monolingual dataset on hate speech in Turkish language from Twitter. There are 1033 hate speech tweets related to gender and sexual orientation in Istanbul Convention dataset, while 1278 hate speech tweets are connected to refugees in Turkey in Refugee dataset. The tweets have been annotated into 5 labels: 'insult', 'exclusion', 'wishing harm', 'threats' and lastly 'not hate'. For Istanbul Convention dataset, the chosen keywords and filtering process were adjusted accordingly to capture hate speech. For Refugee dataset, random sampling was performed to reduce bias during tweets collection. In Istanbul Convention dataset, 36.78\% tweets are related to 'insult', 11.42\% tweets are categorised in 'exclusion', 3.38\% in 'wishing harm', 0.09\% in 'threat' and 48.30\% as 'not hate'. In Refugee dataset, 14.16\% tweets are affiliated to 'insult', 21.67\% to 'exclusion', 3\% to 'wishing harm', 0.5\% to 'threat' and 60.56\% to 'not hate'.\\
\\
\textbf{3. Wich Covid2021 Dataset:}\footnote{https://github.com/mawic/german-abusive-language-covid-19} Wich et al.\cite{wich2021german} constructed the monolingual dataset from Twitter compiled of 4960 tweets in German. Emphasis has been given on hate speech related to COVID-19 and 65 keywords have been used to collect the data. The dataset has been annotated into two classes: 'Abusive' and 'Neutral'. A major portion of the dataset with 3855 tweets(78\%) has been labelled as 'Neutral', while 1105 tweets(22\%) has been labelled as 'Abusive'.\\
\\
\textbf{4. Tulkens Dataset:}\footnote{https://github.com/clips/hades} Tulkens et al.\cite{tulkens2016dictionary} collected a Dutch hate speech monolingual dataset with 6375 social media comments retrieved from two Facebook sites of Belgian  organizations. The dataset has been annotated with 'racist', 'non-racist' and 'invalid' labels. 'Non-racist' label consists of 4943(77.5\%) comments, 'racist' label is comprised of 1088(17.06\%) comments, while 'invalid' label constitutes 344(0.05\%) comments.\\ 
\\
\textbf{5. Sigurbergsson Dataset:} Sigurbergsson et al.\cite{sigurbergsson2019offensive} have gathered the Danish hate speech monolingual dataset from Facebook and Reddit. It constitutes of 3600 comments with 800(22.2\%) comments from Facebook page of Ekstra Bladet, 1400(38.8\%) comments from Danish subreddit r/Denmark and 1400(38.8\%) comments from subreddit r/DANMAG. Hierarchical annotation process has been followed where 3159(87.75\%) are 'not offensive', while 441(12.25\%) comments are 'offensive'. A total of 252 offensive comments are targeted where 95 are towards individuals, 121 towards groups and 36 towards others.\\
\\
\textbf{6. K-MHaS Dataset:}\footnote{https://github.com/adlnlp/K-MHaS} Lee et al.\cite{lee2022k} collected a multilabel monolingual dataset from Korean news comments. It consists of 109692 utterances which have been annotated into multiple labels. The labels consists of 'Politics', 'Race', 'Origin', 'Religion', 'Physical', 'Age', 'Gender', 'Profanity' and 'Not Hate Speech'. While 59615(54.3\%) utterances are 'Not Hate Speech', 50000(45\%) are 'Hate Speech' under different labels.\\
\\
\textbf{7. Bohra Dataset:}\footnote{https://github.com/deepanshu1995/HateSpeech-HindiEnglish-Code-Mixed-Social-Media-Text}Bohra et al.\cite{bohra2018dataset} collected this dataset from Twitter consisting of 4575 tweets in Hindi English code-mixed language. The data has been fetched in json format, which consists of multiple information such as userid, text, user, replies, re-tweets etc. The samples have been annotated as either 'Hate Speech' or 'Normal'. In total 1529(33.4\%) comments have been annotated as hate speech and 3046(66.57\%) comments have been annotated as non-hate.\\ 
\\
\textbf{8. BD-SHS Dataset:}\footnote{https://github.com/naurosromim/hate-speech-dataset-for-Bengali-social-media} Romim et al.\cite{romim2022bd} compiled this Bengali dataset with 50281 comments from multiple social networking sites like YouTube, Twitter, Facebook etc. A hierarchical approach has been obtained for annotation of the samples, where the samples are first identified as hate speech. Prior to that, the target and the type of hate speech is identified. In total, 26125(51.9\%) comments are non-hate while 24156 (48.04\%) comments are hate speech. 
\subsubsection{Multilingual Datasets:} Multiple international series of language research workshops have considered hate speech as a growing area of research. Along with English, these workshops have also paid attention to low-resource languages for hate speech detection. Details of multilingual low-resource datasets available in these workshops is given below and depicted in Table \ref{table:non-E_data_multi}.\\
\\
\textbf{1. HatEval-2019 Dataset:}\footnote{https://github.com/msang/hateval/} Basile et al.\cite{basile2019semeval} compiled this multilingual dataset comprising of 19600 tweets, which consists of 13000 English posts and rest 6600 Spanish posts. These hate tweets mainly targets immigrants and women. While 10509 tweets are misogynistic in nature, 9091 tweets spews hate about immigrants. The data has been annotated in  Offensive Language Identification Dataset (OLID)\cite{zampieri2019predicting} format, where the posts are initially labelled with binary classification. In following level, the hate speech is again specified with annotation whether the hate speech is against an individual or a community and whether the person is displaying aggression or not.\\
\\
\textbf{2. GermEval-2019 Dataset:}\footnote{https://projects.cai.
fbi.h-da.de/iggsa/
} Wiegand et al.\cite{wiegand2018overview} constructed this dataset from Twitter consisting of 8541 German tweets. The tweets are mainly concerned about the refugee crisis in Germany and have been annotated using two labels: 'Offensive' and 'Neutral'. In this dataset, 2890(33.83\%) tweets have been labelled as 'Offensive' and 5651(66.16\%) tweets have been labelled as 'Neutral'.\\
\\
\textbf{3. OffensEval-2020 Dataset:}\footnote{https://sites.google.com/site/offensevalsharedtask/home} Zampieri et al.\cite{zampieri2020semeval} compiled the multilingual dataset with five languages: English, Greek, Danish, Arabic and Turkish. There are 9093037 English tweets which becomes one of the largest datasets. In case of the other low-resource languages, the dataset constitutes 10000 Arabic tweets, 3600 Danish comments from Facebook, Reddit and Ekstra Bladet newspaper site, 10287 Greek tweets and 35000 Turkish tweets.
\begin{table}[!ht]
\caption{Multilingual Datasets in Low-resource Languages}
\label{table:non-E_data_multi}
\centering
\resizebox{\textwidth}{!}{%
\begin{tabular}{|c|c|c|c|c|c|}
\hline
\textbf{Name}  & \textbf{Platform} & \textbf{Language} & \textbf{Size}    & \textbf{Labels}     & \textbf{Reference}                                         \\ \hline 
\begin{tabular}[c]{@{}c@{}}CONAN Hate Speech\\ Dataset - 2019\footnote{https://github.com/marcoguerini/CONAN}\end{tabular}     & Facebook           & \begin{tabular}[c]{@{}c@{}}French,\\Italian\end{tabular}           & 17119 & \begin{tabular}[c]{@{}c@{}}Islamophobic,\\ Non-Islamophobic\end{tabular}  & Chung et al.\cite{chung2019conan}      \\ \hline

\begin{tabular}[c]{@{}c@{}}MLMA Hate\\ Speech  Dataset\\ - 2019\footnote{https://github.com/ HKUST-KnowComp/MLMA\_hate\_speech}\end{tabular}     & Twitter           & \begin{tabular}[c]{@{}c@{}}French,\\Arabic\end{tabular}           & \begin{tabular}[c]{@{}c@{}}4014,\\
3353\end{tabular} & \begin{tabular}[c]{@{}c@{}}Gender,Sexual orientation,\\Religion,Disability\end{tabular}  & \begin{tabular}[c]{@{}c@{}}Ousidhoum et al.\\\cite{ousidhoum2019multilingual} \end{tabular}     \\ \hline

\begin{tabular}[c]{@{}c@{}}HASOC-2019\\\footnote{https://hasocfire.github.io/hasoc/2019/dataset.html}\end{tabular}   & \begin{tabular}[c]{@{}c@{}}Twitter,\\ Facebook\end{tabular}  & \begin{tabular}[c]{@{}c@{}}Hindi,\\German\end{tabular}   & \begin{tabular}[c]{@{}c@{}}5983,\\4669\end{tabular} &  \begin{tabular}[c]{@{}c@{}}Hate, Offensive\\ or Neither, Profane,\\ Targeted or Untargeted \end{tabular}  & Mandl et al.\cite{mandl2019overview}   \\ \hline

\begin{tabular}[c]{@{}c@{}}HASOC-2021\\\footnote{https://hasocfire.github.io/hasoc/2021/dataset.html}\end{tabular}   & \begin{tabular}[c]{@{}c@{}}Twitter,\\ Facebook\end{tabular}  & \begin{tabular}[c]{@{}c@{}}Hindi,\\English,\\Marathi\end{tabular}   & \begin{tabular}[c]{@{}c@{}} 4594,\\3843,\\1874\end{tabular} &  \begin{tabular}[c]{@{}c@{}}Hate, Offensive\\ or Neither, Profane,\\ Targeted or Untargeted \end{tabular}  & Modha et al.\cite{modha2021overview}   \\ \hline
\end{tabular}%
}
\end{table}
\subsection{Multimodal Datasets:} As social media platforms have incorporated multiple mediums of communication, hate speech and offensive language is also not only limited to texts. Audio, video and images have been used to spread hate speech on different platforms. Researchers have started incorporating every modality for hate speech detection. Details of some multimodal hate speech datasets are provided below.
\subsubsection{MMHS150K Dataset:}\footnote{https://gombru.github.io/2019/10/09/MMHS/} Gomez et al.\cite{gomez2020exploring} compiled this dataset from Twitter consisting of 150000 tweets in both textual and visual mode. Tweets containing less than three words and retweets have been filtered out. Tweets containing screenshots of other tweets are also not considered. Tweets have been annotated by Amazon Mechanical Turk with 6 labels: 'no attacks to any community', 'religion based attacks', 'attacks to other communities', 'racist', 'homophobic' and 'sexist'. In this dataset, 112845 (75.23\%) are
not hate tweets, while 36978 (24.65\%) are hate tweets. Among hate tweets, 163 (0.44\%) are religion-based hate, 11925 (32.24\%) are racist, 3870 (10.46\%) are homophobic, 3495 (9.45\%) are sexist and 5811 (15.71\%) are other hate tweets.
\subsubsection{HATEFUL MEMES Dataset:}\footnote{https://hatefulmemeschallenge.com/} Kiela et al.\cite{kiela2020hateful} of Facebook AI have created this multimodal dataset consisting of 10k memes. A text or a meme can be harmless in itself, but when these two are combined it can become hate speech. If the text or the image in itself is non-hate, then it has been defined by the authors as benign confounder. The memes have been annotated into following classes: benign text confounder, benign image confounder, multimodal hate (benign confounders were found for both modalities),
unimodal hate (one or both modalities were already hateful on their own) and random
non-hateful. A Cohen’s kappa score of 67.2\% has been obtained for this dataset. 
\subsubsection{Russia-Ukraine Dataset:}\footnote{https://github.com/Farhan-jafri/Russia-Ukraine} Thapa et al.\cite{thapa2022multi} compiled the multimodal hate speech dataset from Twitter based on the Russia-Ukraine war. The dataset consists of 5680 text-image pairs of tweets and has been labelled with a binary class: 'hate' or 'no-hate'. While a majority of the dataset with 4934 samples(86.86\%) have been annotated as 'no-hate', only 746(13.13\%) samples have been annotated as 'hate'.
\subsubsection{Perifanos Dataset:} Perifanos et al.\cite{perifanos2021multimodal} collected this multimodal dataset from Twitter which consists of approximately 4004 annotated tweets in total. This dataset has been accumulated based on Greek tweets targeting undocumented immigrants and people who have been deported. Here, 1040(25.97\%) have been annotated as  toxic and 2964(74.02\%) have been labelled as non-toxic tweets.
\subsubsection{MUTE Dataset:}\footnote{https://github.com/eftekhar-hossain/MUTE-AACL22} Hossain et al.\cite{hossain2022mute} accumulated a multimodal hateful memes dataset with captions in Bengali or code-mixed(Bengali+English) language. This dataset consists of 4158 memes which has been labelled as either Hate or Not-Hate. Memes have been collected manually from multiple social media platforms such as Twitter, Facebook and Instagram. Keywords such as Bengali Memes, Bangla Funny Memes etc have been used during the manual search. While, 2572 (61.85\%) samples have been annotated as Not-Hate, 1586 (38.14\%) samples have been annotated as Hate.
\section{Automatic Hate Speech Detection for Different Languages}
\begin{figure}[htbp]
\centerline{\includegraphics[height=6cm,width=0.75\columnwidth]{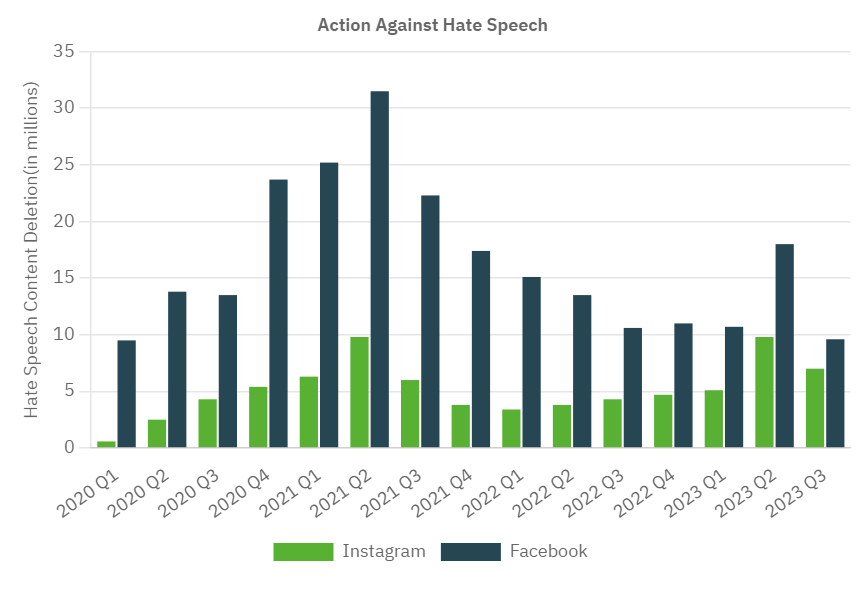}}
\vspace{-0.3cm}
\caption{Action Taken Against Hate Posts on Meta Platforms\cite{statistafb}\cite{statistaig}}
\label{pic:hatecount}
\end{figure}
\begin{figure}
     \centering
     \begin{subfigure}[b]{0.47\columnwidth}
         \centering
         \includegraphics[width=7cm,height=5cm]{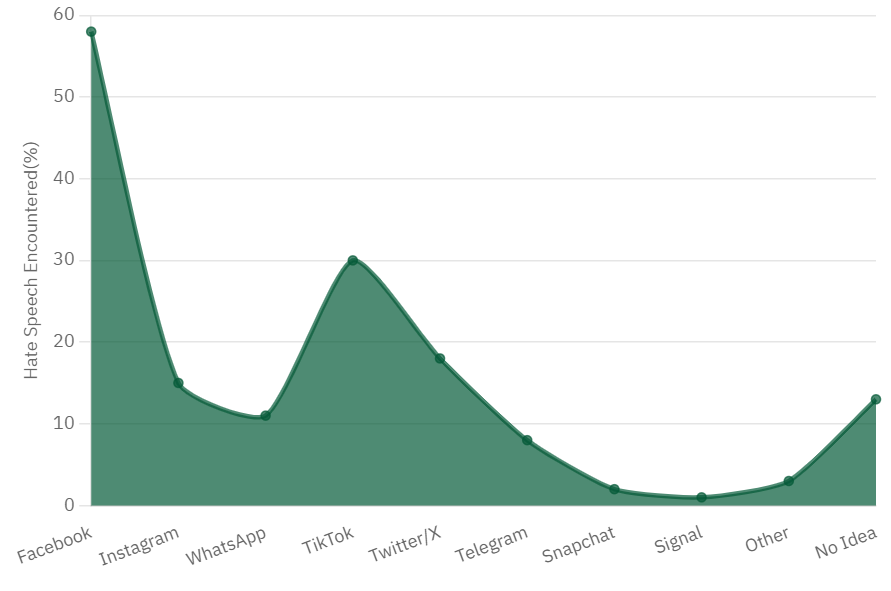}
         \caption{Average Percentage of Hate Speech Encountered on Social Media}
         \label{fig:HateSocial}
     \end{subfigure}
     \hfill
     \begin{subfigure}[b]{0.47\columnwidth}
         \centering
         \includegraphics[width=7cm, height=5cm]{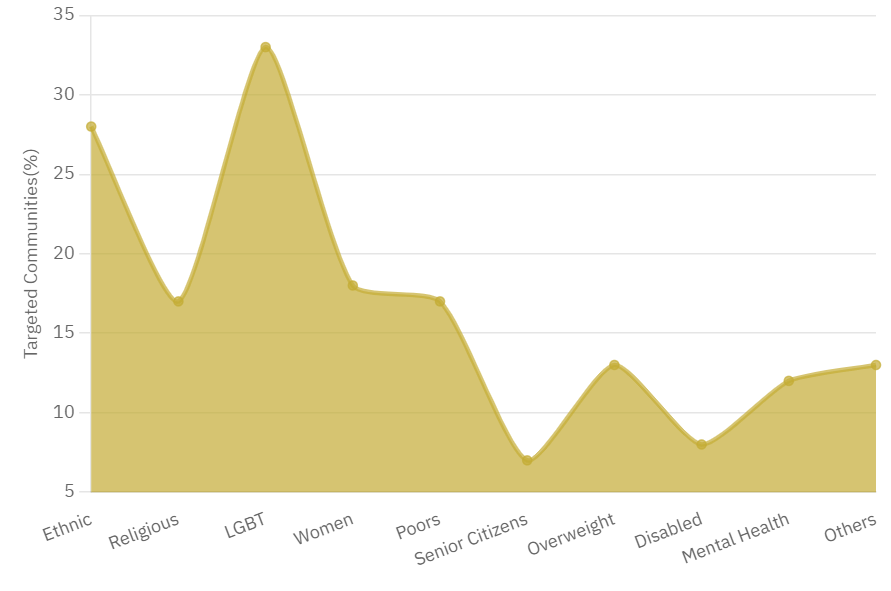}
         \caption{Average Percentage of Targeted Communities}
         \label{fig:Target}
     \end{subfigure}
     \hfill
        \caption{Hate Speech on Social Media across 16 countries\cite{ipsos}}
        \label{fig:IPSOS}
\end{figure}

The rampant utilisation of hate speech has become a bother in various online communities. Social media platforms encourage users to inform and reveal any hate speech encountered online and action is taken against the posts according to company policy. As depicted in Fig. \ref{pic:hatecount}, Meta has released the count of hate speech content deletion on both Facebook and Instagram in different quarters. In a global survey across 16 countries conducted by IPSOS and UNESCO, it has been observed that most hate speech is experienced on Facebook. At the same time, LGBT community is the most targeted community in hate speech. The details of hate speech encountered on various social media platforms has been depicted in Fig. \ref{fig:HateSocial}. The frequently attacked communities on social media are shown in Fig. \ref{fig:Target}.
\subsection{English}
English is the most spoken language globally, which makes it a dominant language for communication. When it comes to social media platforms, websites, most people utilise English while conveying their opinions.  In case of research related to hate speech and abusive words, English is majorly studied because of multiple reasons. The data available from public sites are majorly in English, which leads to considerable number of benchmark datasets. The NLP techniques and parsers developed for English motivated researchers to investigate hate speech detection and conducting comparable studies focusing on the dominant language. Few of them have been explored in this survey.\par
Ghosh et al.\cite{ghosh2022sehc} introduced a English tweets based multi-domain hate speech corpus (MHC) which is compiled using tweets related to diverse domains such as natural disasters, terrorism, technology, and human/drugs trafficking. The dataset contains 10242 samples collected from Twitter having balanced distribution with hate and non hate classes. Using multiple deep learning methods, the authors developed their own classifier named SEHC for binary classification of hate speech. Caselli et al.\cite{caselli2020hatebert} proposes HateBERT which has been re-trained for detection of offensive language in English. A large scale English Reddit comments dataset RAL-E has been used to provide the comparative results with general BERT model, where abusive language inclined HateBERT performed better. Saleh et al.\cite{saleh2023detection} proposed employing domain-specific word embedding and BiLSTM for hate speech detection in English. The hate speech corpus consists of 1048563 sentences depending on which the word embedding is built. Roy et al.\cite{roy2020framework} proposes DCNN model which employs GloVe embedding vector to represent semantics of the posts and eventually hate speech detection. It achieved a precision of 0.97, recall of 0.88 and F1-score of 0.92 for the best case and outperformed the existing models. Rottger et al.\cite{rottger2020hatecheck}proposes HATECHECK which covers model functionalities and compare them with transformer models. Ahluwalia et al.\cite{ahluwalia2018detecting} focuses on misogyny detection with hate tweets directed towards women in tweets that are written in English. Authors have used binary classification using ML techniques for Automatic Misogyny Identification (AMI) at EVALITA2018.
\subsection{Low-Resource Languages}
While there are more than $7000$ spoken languages worldwide, English is considered the standard language internationally. As a result, most research and applications are based on the English language. Supporting toolkits for language preprocessing and cleaning are developed with English as a model language. Research in all the other languages has been initially sparse with researchers providing insignificant attention to these languages. Therefore, except for English, we have considered all the other languages as low-resource languages. This includes European, Asian, African as well as Latin-American languages.
\subsubsection{European Low-Resource Languages:}
Among the different European languages, apart from English, substantial research has been conducted on Western and South-western European languages like Spanish, German, French, etc. as well as Scandinavian languages which include Swedish, Norwegian etc.\\
\textbf{\underline{Spanish}:} Deep learning techniques have been used to detect hate speech in Spanish. Manolescu et al.\cite{manolescu2019tueval} introduce an LSTM model with an embedding layer and character-based representations for Spanish hate speech detection. The Spanish Twitter dataset contains 5000 posts, manually labelled as hate or not, aggressive and targeted. The model achieved an F1-Score of 0.617 for Subtask A and 0.612 for Subtask B. Castillo et al.\cite{castillo2023analyzing} used several datasets  MisoCorpus-2020, DetMis, IberEval2018, including new Twitter dataset named HaSCoSVa-2022 to cover different variants of Spanish. For their transfer learning architecture, the study utilized mBERT and a monolingual Spanish Bert-based language model named BETO. Performance of these models on different variants of Spanish has been compared. The results indicated that models trained for a specific Spanish variant were impacted when multiple variations of the language were ignored. Plaza et al.\cite{plaza2021comparing} uses the HaterNet dataset to evaluate Deep Learning methods and LR, and Transformer mechanism-based pre-trained Language Models that include BERT, BETO, and XLM. The pre-trained Transformer-based language models, provide the finest results. \\
\textbf{\underline{French}:} Automated racist speech detection is implemented using supervised machine learning models. Vanetik et al.\cite{vanetik2022detection} proposes dataset named FTR(French Twitter Racist speech dataset) and performs cross-lingual experiments to test the effectiveness of transfer learning. It confirms that cross-lingual transfer learning is inefficient for detecting racism in French. Battistelli et al.\cite{battistelli2020building} developed a formal model based on cognitive foundations of appraisal theory, linguistic models of modalities in textual data, and various definitions of hate speech. The model identifies hate targets and actions for online communities, emphasizing the significance of context in online hate detection.\\
\textbf{\underline{Dutch}:} Markov et al.\cite{markov2022ensemble} evaluates models using two Dutch social media datasets: LiLaH (Facebook comments) and DALC (tweets). Ensemble method is introduced by combining gradient boosting with transformer-based language models (BERTje and RobBERT) and SVM, along with additional features like hateful words, personal pronouns, and message length. The ensemble approach significantly surpasses independent models in both in-domain and cross-domain hate speech detection settings, providing insights into challenges in Dutch hate speech detection.\\
\textbf{\underline{Polish}:} Ptaszynski et al.\cite{ptaszynski2019results} introduced a dataset for research on cyberbullying in Polish. This dataset has been segregated into binary classification and multi-class classification. This dataset has been collected from Twitter with more than 11k tweets. Korzeniowski et al.\cite{korzeniowski2019exploiting} discusses the challenges faced in hate speech detection using supervised learning due to the scarcity of annotated data. Authors have used unsupervised techniques like fine-tuning of pre-trained BERT and ULMFiT model and have proposed a Tree-based Pipeline Optimization Tool for finding the optimal solution.\\
\textbf{\underline{Austrian German}:} Pachinger et al.\cite{pachinger2024austrotox} introduced a dataset named \textit{Austrotox} which has been collected from an Austrian newspaper named 'DerStandard'. This newspaper reports on multiple topics, both national and international. The authors have filtered out 123108 comments with Austrian dialect on the website of this newspaper. Toxicity score has been assigned to the comments where 873 posts have been categorized as severely toxic, while other posts have lower level of toxicity.\\
\textbf{\underline{German}:} Different NLP techniques have been used for hate speech detection in German. Eder et al.\cite{eder2019lower} proposed an approach that includes a workflow for data-driven acquisition and semantic scaling of a lexicon covering rough, vulgar, or obscene terms. They applied best-worst scaling for rating obscenity and used distributional semantics for automatic lexical enlargement. They have used German slice of Wiktionary, OpenThesaurus, and corpora like CODE ALLTAGS+d email corpus, DORTMUNDER CHAT KORPUS, and FASTTEXT word embeddings based on COMMON CRAWL and WIKIPEDIA. They performed manual editing and utilized distributional semantics methods for lexical expansion. Jaki et al.\cite{jaki2019right} have analyzed more than 50k German hate tweets during the 2017 German federal elections. The authors collected right-wing German Twitter users with subversive profiles. They conduct a comprehensive analysis of characteristics typical of right-wing hate speech, such as the use of derogatory language, dissemination of misinformation, and incitement to violence.\\
\textbf{\underline{Norwegian}:} Hashmi et al.\cite{hashmi2024multi} categorized the dataset into five distinct classes based on the intensity of hate speech, compiled from Facebook, Twitter, and Resset. The paper presents a deep learning approach with GRU and BiLSTM using FastText word embeddings, forming the FAST-RNN model. It also discusses the implementation of multilingual transformer-based models with hyperparameter tuning and generative configuration for Norwegian hate speech detection. The FAST-RNN model outperformed other deep learning models and achieved a high Macro F1-Score.\\
\textbf{\underline{Finnish}:} Jahan et al.\cite{jahan2022finnish} have used a Finnish annotated hate speech dataset collected from the Suomi24 forum. It contains 10.7k sentences with 16.7\% identified as hate speech. The approach includes experiments with the FinBERT pre-trained model and Convolution Neural Network (CNN). They compared FinBERT’s performance with multilingual-BERT (mBERT) and other models using fastText embeddings. Feature engineering strategies like TF-IDF and n-grams were also applied. FinBERT achieved 91.7\% accuracy and 90.8\% F1 score, outperforming all other models including mBERT and CNN with fastText embeddings.\\
\textbf{\underline{Swedish}:} Fernquist et al.\cite{fernquist2019study} proposed automatic hate detection techniques from digital media. Specifically, the language models for Swedish hate speech are fine-tuned and the results are examined based on the pre-trained models. The authors collected labelled data from Swedish online forums like Avpixlat, Flashback, Familjeliv etc and curated 17330 posts. Authors chose comments with entities as hateful comments are often directed towards communities or persons.
\subsubsection{Latin-American Low Resource Languages:} In the Latin American region, substantial work has been organized in Portuguese language used in Brazil. Considerable research on hate speech has been conducted in Latin American Spanish spoken across multiple Latin-American countries.\\
\textbf{\underline{Brazilian Portuguese}:} Multiple datasets like HateBR corpus\cite{vargas2021hatebr} and ToLD-Br\cite{leite2020toxic} have been collected from Brazilian politicians’ Instagram accounts, manually annotated by specialists. HateBR discusses a multi-layer annotation schema for detecting hate speech and offensive language, including classification of offensive and non-offensive, level of offensiveness classification such as high, moderate, and slight offensive, and nine hate speech categories. The baseline experiments on the HateBR corpus achieved an F1-score of 85\%, surpassing the current state-of-the-art for the Portuguese language. ToLD-Br is manually annotated into seven categories and discusses the use of state-of-the-art BERT models for binary classification of toxic comments. Monolingual BERT models trained on Brazilian Portuguese and Multilingual BERT models were used. The paper also explores transfer learning and zero-shot learning using the OLID dataset. The monolingual BERT models achieved a macro-F1 score of up to 76\% using monolingual data in the binary case.\\
\textbf{\underline{Latin-American Spanish}:} Aldana et al.\cite{aldana2021language} presents a method for detecting misogynistic content in texts written in Latin-American Spanish. The approach integrates multiple sources of features to enhance the accuracy of misogyny detection with transformer-based models. Aguirre et al.\cite{aguirre2021problematic} have accumulated 235251 Spanish comments from 200 YouTube videos based on discrimination against Venezuelan refugees in Peru and Ecuador. The comments are mainly categorized as racist, xenophobic and sexist hate speech.
\subsubsection{African Low Resource Languages:} Restricted studies have been conducted for hate speech detection in the case of African languages. Primarily, Amharic and Algerian Arabic have been extensively studied among all the other African languages.\\
\textbf{\underline{Amharic}:} Mossie et al.\cite{mossie2018social} created a corpus of 6120 Amharic Facebook posts and comments. An Apache Spark-based model is proposed that perform feature selection using Word2Vec and TF-IDF, and classification using Naive Bayes and Random Forest. The Word2Vec model with Naïve Bayes classifier achieved the best performance with accuracy and ROC Score of 79.83\% and 0.8305, respectively.\ Tesfaye et al.\cite{tesfaye2020automated} prepared a labelled dataset by curating Amharic posts from certain Facebook pages of activists. The dataset was manually labelled as hate and free speech, pre-processed with data cleaning and normalization techniques. The approach involves developing RNN models using GRU and LSTM. For feature extraction, word n-grams and in case of unique word vector representation, word2vec is used. The LSTM-based RNN model achieved an accuracy of 97.9\% in detecting hate posts. \\
\textbf{\underline{Algerian Arabic}:} Lanasri et al.\cite{lanasri2023hate} created a corpus from Algerian social networks (Facebook, YouTube, and Twitter), containing over 13.5K documents in Algerian dialect written in Arabic, labelled as hateful or non-hateful. The paper evaluates deep learning architectures on the corpus. It discusses the use of Linear Support Vector Classifier (LinearSVC), Gzip + KNN, LSTM \& BiLSTM with Dziri FastText, Dziribert-FT-HEAD, DZiriBERT FT PEFT+LoRA, Multilingual-E5-base FT, sbert-distill-multilingual FT, DzaraShield, and AraT5v2-HateDetect. Results are presented in terms of accuracy, precision, recall, and F1 score, with DzaraShield achieving the highest scores across these metrics.\\
\textbf{\underline{South African Languages}:} Oriola et al.\cite{oriola2020evaluating} evaluated ML techniques using word and character n-gram, syntactic-based features and negative sentiment. They applied ensemble, hyper-parameter optimization and multi-tier meta-learning models using algorithms like SVMs, Random Forest, LRs, and gradient boosting. The best hate speech detection was obtained by SVM with character n-gram having a true positive rate of 0.894. In case of word n-gram,  optimized gradient boosting achieved best results in hate speech detection having true positive rate of 0.867. Multi-tier meta-learning models provided the most consistent and balanced classification performance across different classes. The dataset consists of 21,350 tweets related to prominent individuals, trending issues and 2019 South African elections. Non-English tweets, repeated tweets, and tweets with empty texts were removed. The final corpus contained 14,896 tweets after cleaning and annotation. 
\subsubsection{Asian Low Resource Languages:} Asian languages are the most researched low-resource languages with considerable work in hate speech.  Middle Eastern languages like Arabic, Pashto, Urdu etc. were considered among the initial research of hate speech and abusive words among Asian languages. Later on, research on Southeast Asian languages like Indonesian, Vietnamese etc., East Asian languages like Japanese, Korean etc. and languages of the Indian subcontinent gained traction.\\
\textbf{\underline{Arabic}:} Mubarak et al.\cite{mubarak2017abusive} used two datasets: 175 million tweets obtained from Twitter with language filter set to Arabic for constructing an offensive word list and 1100 manually labelled tweets and 32K user comments removed from an Arabic news site due to policy violations of the site. The approach includes creation of a list of abusive words from the tweets using usual patterns in abusive communications. Based on the words from the abusive list, users are classified into obscene and clean groups. Using list-based methods and considering lexical and syntactical features for detecting abusive language. The results showed that in this case word unigrams are better performing than word bigrams.\\ 
\textbf{\underline{Roman-Urdu and Urdu}:} Akhter et al.\cite{akhter2020automatic} proposed the first offensive dataset of Urdu curated from social media. Character and word-level n-grams techniques for feature extraction have been employed. The study applies 17 classifiers from 7 ML techniques where character n-grams combined with regression-based models performed better for Urdu. LogitBoost and SimpleLogistic outperformed other models, achieving F-measure of 99.2\% on Roman-Urdu and 95.9\% for Urdu datasets. 
Ali et al.\cite{ali2021improving} curated Urdu tweets for sentiment analysis-based hate speech detection. The approach uses Multinomial Naïve Bayes’ (MNB) and SVM. For class imbalance, data sparsity and dimensionality, Variable Global Feature Selection Scheme (VGFSS), dynamic stop words filtering, and Synthetic Minority Optimization Technique (SMOTE) were employed. The results showed that resolving high dimensionality and skewness provided the utmost improvement in the overall performance. \\
\textbf{\underline{Roman-Pashto}:} Khan et al.\cite{khan2023offensive} used a Roman Pashto dataset, created by collecting 60K comments from various social media platforms and manually annotating them. The approach includes: bag-of-words (BoW), term frequency-inverse document frequency (TF-IDF), and sequence integer encoding. The model architectures used are 4 traditional classifiers and a deep sequence model with BiLSTM. The random forest classifier achieved a testing accuracy of 94.07\% using unigrams, bigrams, and trigrams, and 93.90\% with TF-IDF. The highest testing accuracy of 97.21\% was obtained using the BiLSTM model.\\
\textbf{\underline{Japanese}:} Fuchs et al.\cite{fuchs2021normalizing} utilized a Twitter dataset collected and filtered for 19 preselected female politician names with a corpus consisting of 9449645 words. An explorative analysis applying computational corpus linguistic tools and methods, supplemented by qualitative in-depth study is proposed. It combines quantitative-statistical and qualitative-hermeneutic methods to detect and analyze misogynist or sexist hate speech and abusive language on Twitter. Abusive language was present in a significant portion of the negative tweets, with percentages ranging from 33.33\% to 48.6\% for the politicians studied. The analysis also highlighted the use of gendered vocabulary and references to physical appearance in the abuse.  Kim et al.\cite{kim2024impact} collected Twitter data and retweet networks were used to measure adoption thresholds of racist hate speech. Tweets were collected using 15 keywords related to racism against Koreans. The detection of hate speech posts was crucial, and a SVM with TF-IDF values was used for classification.\\ 
\textbf{\underline{Korean}:} Kang et al.\cite{kang2022korean} formed a multilabel Korean hate speech dataset with 35K comments, including 24K online comments, 1.7K sentences from Human-in-the-Loop procedure, 2.2K neutral sentences from Wikipedia, and 7.1K rule-generated neutral sentences. The dataset is designed considering Korean cultural and linguistic context to deal with Western context in English posts. It includes 7 categories of hate speech and employs Krippendorff's Alpha for label accordance. The base model achieved an LRAP accuracy of 0.919.\\
\textbf{\underline{Burmese}:} Nkemelu et al.\cite{nkemelu2022tackling} curated a dataset of 226 Burmese hate speech from Facebook using its CrowdTangle API service. It involves a community-driven process with context experts throughout the ML project pipeline, including scoping the project, assessing hate speech definitions, and working with volunteers to generate data, train, and validate models. Classical machine learning algorithms were employed, with feature combinations of n-grams and term-frequency weighting. The best-performing model was FastText, achieving good precision, recall, and F1-score.\\
\textbf{\underline{Vietnamese}:} Do et al.\cite{do2019hate} used dataset from the VLSP Shared Task 2019: Hate Speech Detection on Social Networks which includes 25431 items. The authors implemented a framework based on the ensemble of Bi-LSTM models for hate speech detection. Word embeddings used were Word2Vec and FastText, with FastText achieving better results. The Bi-LSTM model achieved the best results with a 71.43\% F1-score on the public standard test set of VLSP 2019. Luu et al.\cite{luu2021large} introduced the ViHSD dataset, which is a human-annotated dataset for automatically detecting hate speech on social networks. It contains over 30K comments annotated as Hate, Offensive or Clean. Data was collected from Vietnamese Facebook pages and YouTube videos, with preprocessing to remove name entities for anonymity. They implemented Text-CNN and GRU models with fastText pre-trained word embedding, and transformer models like BERT, XLM-R, and DistilBERT with multilingual pre-training. The BERT model with bert-base-multilingual-cased achieved the best result with 86.88\% accuracy and 62.69\% F1-score on the ViHSD dataset.\\
\textbf{\underline{Indonesian}:} Sutejo et al.\cite{sutejo2018indonesia} develop deep learning models for Indonesian hate speech detection from speech and text. Authors created a new dataset Tram containing posts from various social media such as LineToday, Facebook, YouTube, and Twitter. Both acoustic features for speech and textual features for text were utilized to compare their accuracies. Experiments result depicted that textual features are more efficient than acoustic features for hate speech detection. The best text-based model gained Fl-score of 87.98\% .
\subsubsection{Indian sub-continent Low Resource Languages:} The Indian subcontinent region consists of multiple countries with diverse languages. A miniscule number of languages have been studied regarding hate speech detection. Languages like Bengali, Sinhala, Nepali have large speakers in the different countries of Indian subcontinent and have gained attention from researchers.\\
\textbf{\underline{Sinhala}:} Sandaruwan et al.\cite{sandaruwan2019sinhala} proposed lexicon-based and machine learning-based models for detecting hate speech in Sinhala social media. Corpus-based lexicon generated an accuracy of 76.3\% for offensive, hate, and neutral speech detection. A corpus of 3k comments were curated consisting of texts which are evenly distributed among hate, offensive and neutral speeches. Multinomial Naïve Bayes combined with character tri-gram provided an accuracy of 92.33\% with the best recall value as 0.84.\\
\textbf{\underline{Nepali}:} Niraula et al.\cite{niraula2021offensive} used a dataset consisting of over 15,000 comments and posts from diverse social media platforms such as Facebook, Twitter, YouTube, Nepali Blogs, and News Portals. The dataset was manually annotated with fine-grained labels for
offensive language detection in Nepali. The approach discussed involves supervised machine learning for detecting offensive language. The authors experimented with word and character features, and employed machine learning models. They proposed novel preprocessing methods for Nepali social media text. The results showed that character-based features are extremely useful for classifying offensive languages in Nepali. The Random Forest classifier was chosen for fine-grained classification and achieved F1 scores of 0.87 for Non-Offensive, 0.71 for Other Offensive, 0.45 for Racist, and 0.01 for Sexist categories.\\
\textbf{\underline{Bengali}:} While Bengali as a language is spoken in whole of Bangladesh, it is also the second most spoken language in India. As a result, Bengali has acquired adequate observation in terms of hate speech detection.\par 
ToxLex\_bn by Rashid et al.\cite{rashid2022toxlex_bn}is an exhaustive wordlist curated for detecting toxicity in social media. It consists of 1968 unique bigrams or phrases derived from 2207590 comments. The paper analyzes the Bangla toxic language dataset by developing a toxic wordlist or phrase-list as classifier material. Hossain et al.\cite{hossain2023baad} presented a dataset of 114 slang words and 43 non-slang words with 6100 audio clips. Data was collected from 60 native speakers of slang words and 23 native speakers of non-abusive words. Junaid et al.\cite{junaid2021bangla} created their dataset using Bangla videos from YouTube. It involved machine learning classification methods and deep learning techniques. The logistic regression model achieved the best average accuracy of 96\%. The fine-tuned GRU model achieved an accuracy of 98.89\%, while the fine-tuned LSTM model achieved 86.67\%. 
\begin{table}[htbp]
\begin{tabular}{cccc}
\hline
\textbf{Author}    & \textbf{\begin{tabular}[c]{@{}c@{}}Feature\\ Representation\end{tabular}}   & \textbf{Models}         & \textbf{Metrics}  \\
\hline
\hline
Romim et al.\cite{romim2021hate}  & \begin{tabular}[c]{@{}c@{}}Word2Vec, FastTest, \\ BengFastText\end{tabular} & \begin{tabular}[c]{@{}c@{}}SVM, LSTM, \\ Bi-LSTM\end{tabular} & \begin{tabular}[c]{@{}c@{}}SVM (Acc- 87.5\%,\\ F1-Score - 0.911)\end{tabular} \\
\hline
 Karim et al.\cite{karim2022multimodal}   &  Word Embeddings  &  \begin{tabular}[c]{@{}c@{}}Conv-LSTM, \\ Bi-LSTM, \\m-BERT(cased \& uncased),\\ ResNet-152,\\ DenseNet-161\end{tabular}   &  \begin{tabular}[c]{@{}c@{}}Conv-LSTM (P- 0.79,\\ R-0.78, F1 - 0.78)\\ XLM-RoBERTa (P- 0.82,\\ R- 0.82,F1- 0.82) \\ DenseNet-161 (P- 0.79,\\ R- 0.79, F1- 0.79)\end{tabular}  \\ 
 \hline
Das et al.\cite{das2022hate} &  Linguistic Features  &  \begin{tabular}[c]{@{}c@{}}m-BERT, XLM-Roberta,\\ IndicBERT,  MuRIL\end{tabular}  &  \begin{tabular}[c]{@{}c@{}}MuRIL(Acc- 0.833, \\F1- 0.808)\\ XLM-RoBERTa(Acc- 0.865,\\ F1- 0.810)\end{tabular}\\                   \hline 
Karim et al.\cite{karim2021deephateexplainer} & Linguistic Features & \begin{tabular}[c]{@{}c@{}}m-BERT(cased \& uncased),\\ XLM-Roberta,\\ BiLSTM, \\Conv-LSTM,\\ ML Baselines\end{tabular} & \begin{tabular}[c]{@{}c@{}}XML-RoBERTa (P-0.87,\\ R- 0.87, F1- 0.87)\\Conv-LSTM (P- 0.79,\\ R- 0.78, F1- 0.78)\\GBT (P- 0.71,\\ R- 0.69, F1- 0.68)\end{tabular}\\
\hline
\end{tabular}
\end{table}
Jahan et al.\cite{jahan2022banglahatebert} used a large-scale corpus with abusive and hateful Bengali posts collected from various sources. The authors provided a manually labelled dataset with 15K Bengali hate speech. They considered the existing pre-trained BanglaBERT model and retrained it with 1.5 million offensive posts. The architecture includes Masked-Language Modeling(MLM) and Next Sentence Prediction (NSP) as part of the BERT training strategies. BanglaHateBERT outperformed the corresponding available BERT model in all datasets. Keya et al.\cite{keya2023g} used Bengali offensive text from the social platform (BHSSP), which includes 20,000 posts, comments, and memes from social networks and Bengali news websites. The approach combines BERT architecture and GRU to form G-BERT model. G-BERT achieved an accuracy of 95.56\%, precision of 95.07\%, recall of 93.63\%, and F1-score of 92.15\%. Das et al.\cite{das2021bangla} proposed an encoder–decoder-based ML model, followed by attention mechanism, LSTM, and GRU-based decoders. The attention-based decoder achieved the best accuracy of 77\%. A dataset of 7,425 Bengali comments from various Facebook pages was used. The comments were classified into 7 categories: Hate speech, aggressive comment, religious hatred, ethnical attack, religious comment, political comment, and suicidal comment. Ishmam et al.\cite{ishmam2019hateful} developed ML algorithms and GRU-based deep neural network model. They employed word2vec for word embeddings and compared several ML algorithms. The best performance was achieved by the GRU-based model with 70.10\% accuracy, improving upon the Random Forest model's 52.20\% accuracy. Belal et al.\cite{belal2023interpretable} uses the dataset that consists of 16,073 instances, with 7,585 labeled as Non-toxic and 8,488 as Toxic. The toxic instances are manually labeled in one or more of six classes – vulgar, hate, religious, threat, troll, and insult. Data was accumulated from three sources: Bangla-Abusive-Comment-Dataset, Bengali Hate Speech Dataset, and Bangla Online Comments Dataset. The authors proposed a two-stage deep learning pipeline: a binary classification model (LSTM with BERT Embedding) determines if a comment is toxic and if toxic, a multi-label classifier (CNN-BiLSTM with attention mechanism) categorizes the toxicity type. The binary classifier achieved 89.42\% accuracy, multi-label classifier attained 78.92\% accuracy and a weighted F1-score of 0.86. Banik et al.\cite{banik2019toxicity} used a human-annotated dataset with labels such as toxic, threat, obscene, insult, and racism. The dataset include 10,219 total comments, with 4,255 toxic and 5,964 non-toxic comments. The CNN model included a 1D convolutional layer and global max-pooling, while the LSTM model processed one-hot encoded vectors of length 50. The deep learning-based models, CNN and LSTM, outperformed the other classifiers by a 10\% margin, with CNN achieving the highest accuracy of 95.30\%. Sultana et al.\cite{sultana2023detection} collected a dataset of 5,000 data points from Facebook, Twitter, and YouTube. The dataset is labeled as abusive or not abusive and various ML algorithms are performed for hate speech detection. Sazzed\cite{sazzed2021identifying} compiled two Bengali review corpora consisting of 7,245 comments collected from YouTube, manually annotated into vulgar and non-vulgar categories after preprocessing to exclude English and Romanized Bengali. BiLSTM model yielded the highest recall scores in both datasets. Haque et al.\cite{haque2023multi} proposed a supervised deep learning classifier based on CNN and LSTM for multi-class sentiment analysis. It involved training RNN variants, LSTM, BiLSTM, and BiGRU models, on word embedding. Emon et al.\cite{emon2019deep} evaluate several ML and DL algorithms for identifying abusive content in the Bengali language. It introduces new stemming rules for Bengali to improve algorithm performance. The RNN with LSTM cell outperformed other models, achieving the highest accuracy of 82.20\%.\\
\textit{5.2.5.1  Survey on Mono-Lingual Indic Languages Architecture:}\\ 
Recent years have observed a rise in usage of Indic languages leading to easier accumulation of data from online sources.\\
\textbf{\underline{Hindi}:} Hindi is the most widely used Indian language and a majority of the communications on social media are conducted in Hindi. Kannan et al.\cite{kannan2021hatespeech} used dataset of the HASOC 2021 workshop containing tweets in Hindi annotated as ‘NOT’ and ‘HOF’. The approach is a C-BiGRU model, combining a CNN with a bidirectional RNN. It uses fastText word embeddings and is designed to capture contextual information for binary classification of offensive text. The C-BiGRU model achieved a macro F1 score, accuracy, precision, and recall of 75.04\%, 77.48\%, 74.63\%, and 75.60\% respectively. Bashar et al.\cite{bashar2020qutnocturnal} pretrained word vectors on a collection of relevant tweets and training a CNN model on these vectors. The CNN model achieved an accuracy of 0.82 on the test dataset, with precision, recall, and F1-score for the HOF class being 0.85, 0.74, and 0.79 respectively, and for the NOT class being 0.8, 0.89, and 0.84 respectively. Sreelakshmi et al.\cite{sreelakshmi2020detection} uses Facebook’s pre-trained word embedding library, fastText, for representing data samples. It compares fastText features with word2vec and doc2vec features using SVM-RBF classifier. The dataset consists of 10000 texts equally divided into hate and non-hate classes. Rani et al.\cite{rani2020comparative} used three datasets:(1) 4575 Hindi-English code-mixed annotated tweets in Roman script only, (2) HASOC2019, with 4665 annotated posts collected from Twitter and Facebook (3) Author-created, containing 3367 tweets annotated by them, in both Roman and Devanagari script. They used a character-based CNN model, TF weighting as ML classifiers feature, and no feature engineering or preprocessing for the CNN model. The character-level CNN model outperformed other classifiers. The accuracy for the combined dataset using the CNN model was 0.86, and the micro F1 score was 0.74. Sharma et al.\cite{sharma2024thar} compiled the dataset,"THAR," consisting of 11549 YouTube comments in Hindi-English code-mixed language, annotated for religious discrimination. The paper explores the application of deep learning and transformer-based models. It discusses the use of word embeddings and the evaluation of model performance on the curated dataset. MuRIL outperformed others, achieving macro average and weighted average F1 scores of 0.78 for binary classification and 0.65 and 0.72 for multi-class classification, respectively.\\
\textbf{\underline{Marathi}:} Marathi is considered as the third most spoken language in India. Multiple HASOC datasets have been formed from online communications in Marathi for abusive language detection. Chavan et al.\cite{chavan2022twitter} used the HASOC 2021, HASOC 2022, and MahaHate datasets. The authors have focused on MuRIL, MahaTweetBERT, MahaTweetBERT-Hateful, and MahaBERT models. The MahaTweetBERT model, pre-trained on Marathi tweets and fine-tuned on the combined dataset (HASOC 2021 + HASOC 2022 + MahaHate), achieved an F1 score of 98.43 on the HASOC 2022 test set, providing a new state-of-the-art result on HASOC 2022 / MOLD v2 test set. Gaikwad et al.\cite{gaikwad2021cross} introduced MOLD (Marathi Offensive Language Dataset), the first dataset for offensive language identification in Marathi. The paper discusses ML experiments including zero-shot and transfer learning experiments on cross-lingual transformers from existing data in Bengali, English, and Hindi. Models like SVMs, LSTM, and deep learning models like BERT-m and XLM-Roberta (XLM-R) were used. Transfer learning from Hindi outperformed other methods with a macro F1 score of 0.9401. Weighted F1 scores were also reported. Velankar et al.\cite{velankar2022mono} have used datasets like  HASOC’21 Marathi dataset, L3Cube-MahaSent etc. to present a comparative study of both monolingual BERT and multilingual BERT based models. Ghosh et al.\cite{ghosh2022hate} have examined the effectiveness of mono and multilingual transformer models in Indic languages.\\
\textbf{\underline{Assamese}:} In north-east India, majority speaks Assamese language. Recently, researchers have started giving attention to studying hate speech in Assamese. Ghosh et al.\cite{ghosh2023transformer} annotated Assamese dataset with 4,000 sentences. The approach involves fine-tuning two pre-trained BERT models: mBERT-cased (bert-base-multilingual-cased) and Bangla BERT (sagorsarker/bangla-bert-base), using the Assamese data for hate speech detection. Results are based on weighted F1 scores, precision, recall, and accuracy metrics. The mBERT-cased model achieved an accuracy of 0.63, and the Bangla-BERT model achieved an accuracy of 0.61. They also used the datasets for hate speech detection in Assamese, Bengali, and Bodo languages, collected from YouTube and Facebook comments\cite{ghosh2023annihilate}. Each dataset is binary classified (hate or non-hate). The paper discusses the use of supervised machine learning systems, with a focus on a variant of BERT architecture achieving the best performance. Other systems applied include deep learning and transformer models. The best classifiers for Assamese, Bengali, and Bodo achieved Macro F1 scores of 0.73, 0.77, and 0.85, respectively.\\
\textbf{\underline{Dravidian}:} Dravidian contains the languages spoken in the southern region of India, which consists of Tamil, Telugu, Kannada, Malayalam etc. Among all the Dravidian languages, Tamil has been researched extensively for offensive language detection. Das et al.\cite{das2022data} used datasets in 8 different languages from 14 publicly available sources. The datasets vary in their choice of class labels and are combined into a binary classification task of abusive versus normal posts. The datasets include languages such as Bengali, English, Hindi (Devanagari and code-mixed), Kannada (code-mixed), Malayalam (code-mixed), Marathi, Tamil (code-mixed), and Urdu (actual and code-mixed). It explores various transfer mechanisms for abusive language detection, such as zero-shot learning, few-shot learning, instance transfer, cross-lingual learning, and synthetic transfer. Models like m-BERT and MuRIL are used, which are pre-trained on multiple languages. Chakravarthi et al.\cite{chakravarthi2023offensive} used the DravidianCodeMix dataset, which consists of code-mixed comments on Tamil, Malayalam, and Kannada movie trailers from YouTube. The approach discussed involves a fusion model combining MPNet (Masked and Permuted Network) and CNN for detecting offensive language in low-resource Dravidian languages. The model achieved better offensive language detection results than other baseline models with weighted average F1-scores of 0.85, 0.98, and 0.76 for Tamil, Malayalam, and Kannada respectively. It outperformed the baseline models EWDT and EWODT. Vasantharajan et al.\cite{vasantharajan2022towards} proposes an approach that includes selective translation and transliteration techniques for text
conversion, and extensive experiments with BERT, DistilBERT, XLM-RoBERTa, and CNN-BiLSTM. The ULMFiT model was used with AWD-LSTM architecture and FastAI for fine-tuning. The ULMFiT model yielded the best results with a weighted average F1-score of 0.7346 on the test dataset. The dataset used is a Tamil-English code-mixed dataset from YouTube comments/posts, compared with the Dakshina dataset for out-of-vocabulary words. Subramanian et al.\cite{subramanian2022offensive} compares traditional machine learning techniques with pre-trained multilingual transformer-based models using adapters and fine-tuners for detecting offensive texts. Transformer-based models outperformed machine learning approaches, with adapter-based techniques showing better performance in terms of time and efficiency, especially in low-resource languages like Tamil. The XLM-RoBERTa (Large) model achieved the highest accuracy of 88.5\%. Anbukkarasi et al.\cite{anbukkarasi2023deep} proposes a dataset of 10,000 Tamil-English code-mixed texts collected from Twitter, annotated as hate text/non-hate text. A Bi-LSTM model has been proposed to classify hate and non-hate text in tweets. Hande et al.\cite{hande2020kancmd} proposed the dataset named KanCMD, a multi-task learning dataset for sentiment analysis and offensive language identification in Kannada. It contains comments from YouTube, annotated for sentiment analysis and offensive language detection.The paper presents the dataset statistics and discusses the results of experiments with several machine learning algorithms. The results are evaluated in terms of precision, recall, and F1-score. Using this model Roy et al.\cite{roy2022hate} have proposed an ensemble model for detection of hate speech and offensive language in Dravidian languages. The model has obtained a weighted F1-score of 0.802 and 0.933 for Malayalam and Tamil code-mixed datasets respectively.\\
\\
\textit{5.2.5.2  Survey on Multi-Lingual Indic Languages Architecture:}\\
Deep learning based methods like LSTM variants, mostly transformer language models such as BERT variants, XLM-RoBERTa have been widely used for detection of hate speech in Indian languages. Khanuja et al.\cite{khanuja2021muril} have  designed Multilingual Representations for Indian Languages (MuRIL), a multilingual language model for Indic languages and it has been specifically trained large amounts of Indian texts. \par
 Khurana et al.\cite{khurana2022animojity} developed a novel model called AniMOJity, which incorporates transfer learning and  XLM-RoBERTa along with finetuning it with additional layers.The dataset used is the Moj Multilingual Abusive Comment Identification dataset. It includes comments in thirteen Indian regional languages. 
\begin{table}[htbp]
\caption{Multilingual Research on HASOC Datasets}
\label{table:HASOC}
\centering
\begin{tabular}{cccc}
\hline
\textbf{Authors} & \textbf{Languages} & \textbf{Models}      & \textbf{Results} \\
\hline
\hline
Bhatia et al.\cite{bhatia2021one}      & \begin{tabular}[c]{@{}c@{}}HASOC 2021:\\ English, Hindi, Marathi\end{tabular}                      & \begin{tabular}[c]{@{}c@{}}XLM-RoBERTa,\\ mBERT,DistilmBERT\end{tabular}                                      & \begin{tabular}[c]{@{}c@{}}Macro F1 Scores:\\ 0.7996, 0.7748, 0.8651\end{tabular}    \\
\hline
Velankar et al.\cite{velankar2021hate} & \begin{tabular}[c]{@{}c@{}}HASOC 2021:\\ English, Hindi, Marathi\end{tabular}                      & \begin{tabular}[c]{@{}c@{}}mBERT, IndicBERT, \\ monolingual RoBERTa,\\ CNN, LSTM with\\ FastText\end{tabular} & \begin{tabular}[c]{@{}c@{}}Macro F1 Scores:\\ Marathi- 0.869\\ Hindi- 0.763\end{tabular}  \\
\hline
Narayan et al.\cite{narayan2023hate}   & \begin{tabular}[c]{@{}c@{}}HASOC 2023:\\ Bengali, Assamese, Bodo,\\ Sinhala, Gujarati\end{tabular}   & \begin{tabular}[c]{@{}c@{}}XLM-R,\\ LSTM models\end{tabular}                                                  & \begin{tabular}[c]{@{}c@{}}Macro F1 Scores:\\ Bengali- 0.67027, \\ Assamese- 0.70525 \\ Bodo- 0.83009 \\ Sinhala- 0.83493\\ Gujarati- 0.76601\end{tabular} \\
\hline
Joshi et al.\cite{joshi2023harnessing} & \begin{tabular}[c]{@{}c@{}}HASOC 2023:\\ Bengali, Assamese, Bodo,\\ Sinhala, Gujarati\end{tabular} & \begin{tabular}[c]{@{}c@{}}SBERT,\\ Pre-trained BERT\end{tabular}  & \begin{tabular}[c]{@{}c@{}}Macro F1:\\ Bengali- 0.7703\\ Gujarati- 0.7324\\ Assamese- 0.7065\end{tabular} \\
\hline
\end{tabular}
\end{table}
Vashistha et al.\cite{vashistha2020online} analyzed six publicly available datasets in English and Hindi and combined them into a single homogeneous dataset. They classified the data into three classes: abusive, hateful, or neither. The proposed multilingual model architecture includes logistic regression, CNN-LSTM, and BERT-based networks. Das et al.\cite{das2024low} created a benchmark dataset of 5062 abusive speech/counterspeech pairs, with 2460 pairs in Bengali and 2602 pairs in Hindi. The authors experimented with several transformer-based baseline models for counterspeech generation, including GPT2, MT5, BLOOM, and ChatGPT. They evaluated several interlingual mechanisms and observed that the monolingual setting yielded the best performance. Gupta et al.\cite{gupta2022multilingual} introduced the MACD dataset, a large-scale, human-annotated, multilingual abuse detection dataset sourced from ShareChat. It contains 150K textual comments in 5 Indic languages, with 74K abusive and 77K non-abusive comments. The paper presents AbuseXLMR, a model pre-trained on 5M+ social media comments in 15+ Indic languages. It is based on the XLM-R model and is adapted to the social media domain to bridge the domain gap. Jhaveri et al.\cite{jhaveri2022toxicity}t used the dataset provided by ShareChat/Moj in IIIT-D Multilingual Abusive Comment Identification challenge, containing 665k+ training samples and 74.3k+ test samples in 13 Indic languages labeled as Abusive or Not. The approach included leveraging multilingual transformer-based pre-trained and fine-tuned models such as XLM-RoBERTa and MuRIL. Parikh et al.\cite{parikh2023regional} used dataset which consists of comments in Gujarati, Hindi, English, Marathi, and Punjabi. It contains 7500 non-toxic and 7495 toxic comments. LSTM, CNN and DistilBERT Embedding are used and fine-tuned on the multilingual dataset. Various implementations on the multiple HASOC datasets is depicted in Table \ref{table:HASOC}.
\section{Research Challenges and Opportunities}
Automatically spotting offensive or hateful speech is challenging for plenty of factors, especially while using social media. Most of the challenges are closely associated with the shortcomings of keyword-based techniques. In this section, we highlight significant challenges derived from survey literature of hate speech recognition. 

\noindent\textbf{Culture and Language Constraint} \\
Hate speech is both cultural and linguistic specific. Therefore, cultural specific responses play a very crucial role in hate speech recognition. As Spanish is a widely spoken language around multiple countries, the same words in Spanish can carry different implications based on the country in which they are spoken\cite{castillo2023analyzing}. Moreover, several stereotypes corresponding to the historical events of their regions may apply to the same population \cite{laurent2020project}. Language shifts promptly, especially among younger generations who interact with social media on a daily basis, requiring continuous research on Hate speech datasets. As of now, hate content is being eliminated both manually and automatically from the internet based on the reports submitted by users. Nevertheless, hate speech content spreaders are continuously looking for new methods of getting around and beyond any limitations set by the system. As an illustration, some users upload hate speech as photos for purpose of getting around some automatic hate speech detection systems. Despite the possibility that image to text may address a specific issue, many complications exist due to the constraints around this type of communication and the state of automatic high-spectrum identification. Moreover, updating the linguistic structure may include additional challenges. Often hate speech spreaders use abbreviations, combine multiple languages into the same sentence, use various phonetics where a slight change can convert a simple word into an abusive word. Sometimes, hate spreaders use sarcasm to mock and bully others on social platforms, which is extremely challenging to detect.

\noindent\textbf{Dataset} \\
In the field of research, no datasets are considered to be suitable for automatic hate speech identification task. According to the task demand and the comprehension level, the annotators annotate the dataset in various techniques. Moreover, the utilization of crowd-sourcing also raises concerns about the expertise level of annotators. Hate speech has no clear boundaries of definition and multiple concepts of detrimental languages intermingle with one another. A precise label definition is essential to distinguish hate speech from other categories of harmful languages. Additionally, the information can target a wider spectrum of specific hate speech characteristics such as personal abuse, sexism, racism, cyberbullying, and trolling.
This issue can be resolved in two ways: (1) Multi-labelling Method, and (2) Hierarchical Approach. In multi-labelling method, the labels of sexism and racism may indicate ambiguity. Whereas, the sub-types of Hate Speech along with aggression is described using hierarchical approach. Additionally, many utterances are not intrinsically offensive, but they may be within appropriate circumstances \cite{ullmann2020quarantining}. However, additionally in the case of abusive language, the offense can differ depending on various audience members and users.

\noindent\textbf{Data Collection} \\
Possible biases may arise from various decisions made in the course of collecting data. Primarily, style and content can be significantly effected by the data source. Crawling involves collecting texts from a particular platform which determines its length, style, and themes. These writings mainly represents what the websites offer users and therefore are not very generalized. When utilizing crowdsourcing and nichesourcing, annotators' levels of sensitivity to hate differ based on country of origin, their personal experiences, and membership in the targeted minority \cite{lee2023crehate}. Furthermore, bias can also originate from other sources such as the targets of language, hate, historical, and geographical context of the collected data. Sometimes data collected from social media requires ID retrieval, which is effective when the social media posts are live. But later on receiving reports from other users, platforms often remove these posts from their sites. In such a scenario, the dataset gets reduced and hinders the training of models. 

\noindent\textbf{Open Platform} \\
Indeed, there exists many open source initiatives pertaining to hate speech recognition. However, only few well known publications have made their source code publicly available. Only 53 of the 1039 projects of Github are regularly updated and forked, raising concerns about the remaining projects' source code quality and functionality. This research field can advance more rapidly if it incorporates enhanced code sharing along with comprehensive documentation, efficient algorithms, and publicly available datasets.

\noindent\textbf{Multimodal and Multilingual Characteristic}\\
As the dominant language around the world, resources and approaches in English are comparatively abundant. Presently, multiple datasets are being accumulated\cite{awal2023model} in multiple languages and significant advances have been achieved in low-resource languages. Various datasets have been generated consisting of textual data, but the evolving social media space have incorporated different modalities for spread of hateful content. Images and texts which individually may be harmless, when combined can generate hate speech. Audio-visual mediums can also be used to perpetuate harmful language. Minimal focus has been provided to the multimodal aspect of hate speech detection. Researchers are nowadays concentrating on the multimodal properties\cite{arya2024multimodal} for hate speech.\par
There is a lot of potential for future research opportunities in the domain of hate speech detection. Following are some of the research sections which need attention and improvement.
\begin{itemize}
    \item Improving research in the understanding of language nuances and cultural references. Cross-language detection with the incorporation of various dialects can improve hate speech detection models. 
    \item Merging multiple modalities and promotion of interdisciplinary collaboration among language experts, sociologists, and computer researchers to strengthen detection algorithms.
    \item Detection of hate speech in real-time and monitoring the systems in live social media platforms. Interactions with different communities to understand their perception of hate speech and create models that are culturally and linguistically responsive.
    \item Establishment of transparent ethical frameworks and policies along with moderation tools for building trust among the users.
    \item Investigating the behaviour of users who are involved in hate speech, can assist in developing robust algorithms for targeted interventions and preventive measures. 
    \item Analyzing emotion and sentiment to distinguish between extremely subjective language and hate speech.
    \item Enhancing accuracy detection and interpreting neighbouring contexts by utilizing community feedback and crowd-sourced data.
\end{itemize}
\section{Conclusion}
The continual presence of the internet has facilitated the integration of social media platforms in everyday life. Moreover, people display hostility for a variety of reasons. Perusing this large volume of data for automatic harmful language detection is challenging. In this survey, we have presented an extensive overview of advances obtained in online hate speech detection techniques to date. We have discussed the overlying hate speech concepts of cyberbullying, offensive and abusive language. Initially, the publicly available datasets in both English and other low-resource languages have been detailed. We have identified prevailing reviews, surveys and implementation approaches on hate speech. Particular attention has been provided to hate speech detection in low-resource languages and the techniques applied. Primitively, TF-IDF and other machine learning techniques were used which have been replaced by FastText, word2Vec, GloVe word embeddings combined with deep learning approaches like CNN, RNN and BERT. Multiple research gaps, limitations and challenges have been elaborated with prospective future studies are elaborated. This survey will definitely be useful for researchers pursuing analysis in hate speech detection in multiple low-resource languages. 
\section*{COMPLIANCE WITH ETHICAL STANDARDS}
The authors declare no conflict of interest. No human or animal subjects have been studied in this
article. This article surveyed social network analysis applications.
	
	
	\bibliography{main}
	\appendix

\end{document}